%% file: main.tex
\documentclass[twoside,journal]{IEEEtran}

\usepackage{amsmath,amsfonts,amssymb}
\usepackage{enumitem}
\usepackage{graphicx}
\usepackage{silence}
\usepackage[colorlinks=false]{hyperref}
\usepackage[utf8]{inputenc}
\usepackage[USenglish]{babel}
\usepackage[T1]{fontenc}
\usepackage[keeplastbox]{flushend} %balance the 2 column bibliography
\usepackage[normalem]{ulem} %\sout to strike out
\usepackage[dvipsnames]{xcolor}
\usepackage{microtype}

\makeatletter
\def\markboth#1#2{%
  \def\leftmark{\@IEEEcompsoconly{\sffamily}#1}%
  \def\rightmark{\@IEEEcompsoconly{\sffamily}#2}}
\makeatother

\title{DAC-h3: A Proactive Robot Cognitive Architecture to Acquire and Express Knowledge About the World and the Self}

\author{Clément Moulin-Frier*, Tobias Fischer*, Maxime Petit, Grégoire Pointeau, Jordi-Ysard Puigbo, Ugo Pattacini, Sock Ching Low, Daniel Camilleri, Phuong Nguyen, Matej Hoffmann, Hyung Jin Chang, Martina Zambelli, Anne-Laure Mealier, Andreas Damianou, Giorgio Metta, Tony J. Prescott, Yiannis Demiris, Peter Ford Dominey, and Paul F. M. J. Verschure %
\thanks{Manuscript received December 31, 2016; revised July 25, 2017; accepted August 09, 2017. The research leading to these results has received funding from the European Research Council under the  European Union's Seventh Framework Programme (FP/2007-2013) / ERC Grant Agreement n. FP7-ICT-612139 (What You Say Is What You Did project), as well as the ERC’s CDAC project: Role of Consciousness in Adaptive Behavior (ERC-2013-ADG 341196). M. Hoffmann was supported by the Czech Science Foundation under Project GA17-15697Y. P. Nguyen received funding from ERC's H2020 grant agreement No. 642667 (SECURE). T. Prescott and D. Camilleri received support from the EU Seventh Framework Programme as part of the  Human Brain (HBP-SGA1, 720270) project.}
\thanks{C. Moulin-Frier, J.-Y. Puigbo, S. C. Low, and P. F. M. J. Verschure are with the Laboratory for Synthetic, Perceptive, Emotive and Cognitive Systems, Universitat Pompeu Fabra, 08002 Barcelona, Spain. P. F. M. J. Verschure is also with Institute for Bioengineering of Catalonia (IBEC), The Barcelona Institute of Science and Technology (BIST) and Institució Catalana de Recerca i Estudis Avançats (ICREA), Barcelona, Spain.}
\thanks{T. Fischer, M. Petit, H. J. Chang, M. Zambelli, and Y. Demiris are with the Personal Robotics Laboratory, Department of Electrical and Electronic Engineering, Imperial College London, SW7 2AZ, U.K.}
\thanks{G. Pointeau, A.-L. Mealier, and P. F. Dominey are with the Robot Cognition Laboratory of the INSERM U846 Stem Cell and Brain Research Institute, Bron 69675, France.}
\thanks{U. Pattacini, P. Nguyen, M. Hoffmann, and G. Metta are with the Italian Institute of Technology, iCub Facility, Via Morego 30, Genova, Italy. M. Hoffmann is also with the Department of Cybernetics, Faculty of Electrical Engineering, Czech Technical University in Prague.}
\thanks{D. Camilleri, A. Damianou, and T. J. Prescott are with the Department of Computer Science, University of Sheffield, U.K. A. Damianou is now at Amazon.com.}
\thanks{Digital Object Identifier \href{http://doi.org/10.1109/TCDS.2017.2754143}{10.1109/TCDS.2017.2754143}}
\thanks{}
\thanks{*C. Moulin-Frier and T. Fischer contributed equally to this work.}
}

\begin{document} 
\bstctlcite{IEEEexample:BSTcontrol}
\markboth{Preprint version; final version available at http://ieeexplore.ieee.org/ \\
IEEE Transactions on Cognitive and Developmental Systems (Accepted) \\
DOI: 10.1109/TCDS.2017.2754143}%
{\uppercase{C. Moulin-Frier, T. Fischer} \textit{et al.:} \uppercase{DAC-h3: A Proactive Robot Cognitive Architecture}}
\maketitle

\input{0_abstract}
\input{1_introduction}

\input{2_relatedworks}
\input{3_cog_arch}

%\input{4_0_cog_arch_for_hri}
%\input{4_1_world}
%	\input{4_2_self}
%\input{4_3_actions}
%\input{4_4_contextual}
\input{4_experiments}

\input{5_conclusion_futureworks}
%\input{7_conclusion}

%\section*{Acknowledgment}

% References
\bibliography{report}
\bibliographystyle{IEEEtran}

\end{document}

%% file: 0_abstract.tex
\begin{abstract}
This paper introduces a cognitive architecture for a humanoid robot to engage in a proactive, mixed-initiative exploration and manipulation of its environment, where the initiative can originate from both the human and the robot. The framework, based on a biologically-grounded theory of the brain and mind, integrates a reactive interaction engine, a number of state-of-the art perceptual and motor learning algorithms, as well as planning abilities and an autobiographical memory. The architecture as a whole drives the robot behavior to solve the symbol grounding problem, acquire language capabilities, execute goal-oriented behavior, and express a verbal narrative of its own experience in the world. 
We validate our approach in human-robot interaction experiments with the iCub humanoid robot, showing that the proposed cognitive architecture can be applied in real time within a realistic scenario and that it can be used with naive users.
\end{abstract}


%% file: 1_introduction.tex
\section{Introduction}
\label{sec:intro}
%
%\cmf{I've rewritten the Intro in an attempt to better fit with the SI CfP. As usual I've tried to reuse most of the existing text. Both versions (new and old) below.}
%
%\cmf{New version}
%
\IEEEPARstart{T}{he} so-called \emph{Symbol Grounding Problem} (SGP,~\cite{Harnad1990,Coradeschi2003, taniguchi2016symbol}) refers to the way in which a cognitive agent forms an internal and unified representation of an external word referent from the continuous flow of low-level sensorimotor data generated by its interaction with the environment. In this paper, we focus on solving the SGP in the context of human-robot interaction (HRI), where a humanoid iCub robot~\cite{metta2010icub} acquires and expresses knowledge about the world by interacting with a human partner. Solving the SGP is of particular relevance in HRI, where a repertoire of shared symbolic units forms the basis of an efficient linguistic communication channel between the robot and the human.
To solve the SGP, several questions should be addressed:
\begin{itemize}
\item How are unified symbolic representations of external referents acquired from the multimodal information collected by the agent (e.g., visual, tactile, motor)? This is referred to as the \emph{Physical SGP}~\cite{vogt2002physical,vogt2007social}.
\item How to acquire a shared lexicon grounded in the sensorimotor interactions between two (or more) agents? This is referred to as the \emph{Social SGP}~\cite{vogt2007social,cangelosi2006grounding}.
\item How is this lexicon then used for communication and collective goal-oriented behavior? This refers to the functional role of physical and social symbol grounding.
\end{itemize}

This paper addresses these questions by proposing a complete cognitive architecture for HRI and demonstrating its abilities on an iCub robot. % that solves the SGP (both its physical and social aspects), and in turn acquires complex goal-oriented behavior and the ability to verbally express a self-centered narrative. 
%To address these questions, we propose an embodied and integrated cognitive architecture based on an established theory of the brain and mind, which is able to solve the three aforementioned requirements. 
Our architecture, called \emph{DAC-h3}, builds upon our previous research projects in conceiving biologically grounded cognitive architectures for humanoid robots based on the Distributed Adaptive Control theory of mind and brain (DAC, presented in the next section). In \cite{lallee2015towards} we proposed an integrated architecture for generating a socially competent humanoid robot, demonstrating that gaze, eye contact and utilitarian emotions play an essential role in the psychological validity or social salience of HRI (DAC-h1). In \cite{vouloutsi2016towards}, we introduced a unified robot architecture, an innovative Synthetic Tutor Assistant (STA) embodied in a humanoid robot whose goal is to interactively guide learners in a science-based learning paradigm through rich multimodal interactions (DAC-h2).

\emph{DAC-h3} is based on a developmental bootstrapping process where the robot is endowed with an intrinsic motivation to act and relate to the world in interaction with social peers. Levinson \cite{levinson2006human} refers to this process as the \emph{human interaction engine}: a set of capabilities including looking at objects of interest and interaction partners, pointing to these entities \cite{liszkowski2004twelve}, demonstrating curiosity as a desire to acquire knowledge \cite{berlyne1954theory} and showing, telling and sharing this knowledge with others \cite{liszkowski2004twelve,tomasello2005understanding}. These are also coherent with the \emph{desiderata for developmental cognitive architectures} proposed in \cite{Vernon2016} stating that a cognitive architecture's value system should manifest both exploratory and social motives, reflecting the psychology of development defended by Piaget \cite{Piaget1954} and Vygotsky \cite{Vygotsky1978}. 

This interaction engine drives the robot to proactively control its own acquisition and expression of knowledge, favoring the grounding of acquired symbols by learning multimodal representations of \emph{entities} through interaction with a human partner. In \emph{DAC-h3}, an entity refers to an internal or external referent: it can be either an object, an agent, an action, or a body part. In turn, the acquired multimodal and linguistic representations of entities are recruited in goal-oriented behavior and form the basis of a persistent concept of self through the development of an autobiographical memory and the expression of a verbal narrative. 

%\cmf{I've removed the detailed answers to each of the above questions from the Intro. The corresponding text is now integrated in section II-F}

We validate the proposed architecture following a human-robot interaction scenario where the robot has to learn concepts related to its own body and its vicinity in a proactive manner and express those concepts in goal-oriented behavior. We show a complete implementation running in real-time on the iCub humanoid robot. The interaction depends on the internal dynamics of the architecture, the properties of the environment and the behavior of the human. We analyze a typical interaction in detail and provide videos showing the robustness of our system in various environments (\url{https://github.com/robotology/wysiwyd}). Our results show that the architecture autonomously drives the iCub to acquire a number of concepts about the present entities (objects, humans, and body parts), whilst proactively maintaining the interaction with a human and recruiting those concepts to express more complex goal-oriented behavior. We also run experiments with naive subjects in order to test the effect of the robot's proactivity level on the interaction.

In Section \ref{sec:related_works} we position the current contribution with respect to related works in the field and rely on this analysis to emphasize the specificity of our approach. %Then we present the theoretical basis of the proposed cognitive architecture in Section \ref{sec:dac}. 
Our main contribution is described in Section \ref{sec:wrdac} and consists in the proposal and implementation of an embodied and integrated cognitive architecture for the acquisition of multimodal information about external word referents, as well as a context-dependent lexicon shared with a human partner and used in goal-directed behavior and verbal narrative generation. The experimental validation of our approach on an iCub robot is provided in Section \ref{sec:results}, followed by a discussion in Section~\ref{sec:conclusion}.

%% file: 2_relatedworks.tex
\section{Related Works and Principles of the Proposed Architecture}
\label{sec:related_works}

%\cmf{It would be useful is someone could check if more references from the old version (at the end of this section) could be included in this new version, Those from the list below should be, but it worth to check as well.}
%\cmf{We need references in those domains:
%\begin{itemize}
%\item Typical example of a functional architecture (see functional vs biologically-inspired distinction in Fong and al., 2003): \cite{pineau2003towards,nourbakhsh1999affective, di2013autonomous};
%\item Typical example of a biologically-inspired arch: \cite{adams2000humanoid,dautenhahn1999bringing};
%\item Typical example of a bottom-up, top-down and hybrid architectures for social robot (we have some from the HRI reviews)
%\item  Works solving the SGP  \cite{marocco2010grounding} (including bottom-up ala Steels), \tf{From one of the topic editors: \cite{Taniguchi2016,taniguchi2016symbol}}
%\item  Works dealing with language acquisition (mostly the EU projects I think): \cite{marocco2010grounding} \cite{steels2008symbol} \tf{Very recent from one of the topic editors in TCDS: \cite{Taniguchi2016TCDS}}
%\item  Works dealing with life-long learning and/or narrative \tf{We can use our TCDS paper: \cite{Petit2016TCDS,huang2016modular}}
%\item  Works dealing with autonomous exploration / curiosity / mixed initiative \cite{Schrempf2005,Lutkebohle2009} (I've some + Rachid Alami from Max?) \cite{steunebrink2013resource}
%\end{itemize}
%}

Designing a cognitive robot that is able to solve the SGP requires a set of heterogeneous challenges to be addressed. First, the robot has to be driven by a cognitive architecture bridging the gap between low-level reactive control and symbolic knowledge processing. Second, it needs to interact with its environment, including social partners, in a way that facilitates the acquisition of symbolic knowledge. Third, it needs to actively maintain engagement with the social partners for a fluent interaction. Finally, the acquired symbols need to be used in high-level cognitive functions dealing with linguistic communication and autobiographical memory.

In this section, we review related works on each of these topics along with a brief overview of the solution adopted by the DAC-h3 architecture.

\subsection{Functionally-driven vs. biologically-inspired approaches in social robotics}
\label{sec:functionally-biologically}

The methods used to conceive socially interactive robots derive predominantly from two approaches \cite{Fong2003}. Functionally-designed approaches are based on reverse engineering methods, assuming that a deep understanding of how the mind operates is not a requirement for conceiving socially competent robots (e.g. \cite{pineau2003towards,nourbakhsh1999affective, di2013autonomous}), whilst biologically-inspired robots are based on theories of natural and social sciences and expect two main advantages of constraining cognitive models by biological knowledge: to conceive robots that are more understandable to humans, as they reason using similar principles, and to provide an efficient experimental benchmark from which the underlying theories of learning can be confronted, tested and refined (e.g. \cite{adams2000humanoid,dautenhahn1999bringing,Demiris2014}). One specific approach used by Demiris and colleagues for the mirror neuron system is decomposing computational models implemented on robots into \emph{brain operating principles} which can then be linked and compared to neuroimaging and neurophysiological data \cite{Demiris2014}.

The proposed \emph{DAC-h3} cognitive architecture takes advantage of both methods. It is based on an established biologically-grounded cognitive architecture of the brain and the mind (the DAC theory, presented below) that is adapted for the HRI domain. However, while the global structure of the architecture is constrained by biology, the implementation of specific modules can be driven by their functionality, i.e. using state-of-the-art methods from machine learning that are powerful at implementing particular functions without being directly constrained by biological knowledge. 

\subsection{Cognitive architectures and the SGP}
\label{sec:cogarch-sgp}
Another distinction in approaches for conceiving social robots, which is of particular relevance for addressing the SGP, reflects a divergence from the more general field of cognitive architectures (or \textit{unified theories of cognition} \cite{Newell1990}). Historically, two opposing approaches have been proposed to formalize how cognitive functions arise in an individual agent from the interaction of interconnected information processing modules in a cognitive architecture. Top-down approaches rely on a symbolic representation of a task, which has to be decomposed recursively into simpler ones to be executed by the agent. These rely principally on methods from symbolic artificial intelligence (from the General Problem Solver \cite{Newell1959} to Soar \cite{Laird1987} or ACT-R \cite{Anderson1983}). Although relatively powerful at solving abstract symbolic problems, top-down architectures are not able to solve the SGP \textit{per se} because they presuppose the existence of symbols. Thus they are not suitable for addressing the problem of how these symbols can acquired from low-level sensorimotor signals. The alternative, bottom-up approaches instead implement behavior without relying on complex knowledge representation and reasoning. This is typically the case in behavior-based robotics \cite{Brooks1991}, emphasizing lower-level sensory-motor control loops as a starting point of behavioral complexity as in the Subsumption architecture \cite{Brooks1986}. These approaches are not suitable to solve the SGP either because they do not consider symbolic representation as a necessary component of cognition (referred as \textit{intelligence without representation} in \cite{Brooks1991}). 

Interestingly, this historical distinction between top-down representation based and bottom-up behavior based approaches still holds in the domain of social robotics \cite{Miklosi2012, Dautenhahn2007}. Representation based approaches rely on the modeling of psychological aspects of social cognition (e.g. \cite{Carruthers1998}), whereas behavior based approaches emphasize the role of embodiment and reactive control to enable the dynamic coupling of agents \cite{DiPaolo2012}. 
%However none of these approaches is able to solve the Symbol Grounded Problem \textit{per se}. The reason is that top-down approaches presuppose the existence of symbols and therefore are not suitable for addressing how they are formed from low-level sensorimotor signals; whereas bottom-up approaches do not consider symbolic representation as a necessary component of cognition \cite{Brooks1991}. 
Solving the SGP, both in its physical and social aspects, therefore requires the integration of bottom-up processes for acquiring and grounding symbols in the physical interaction with the (social) environment and top-down processes for taking advantage of the abstraction, reasoning and communication abilities provided by the acquired symbol system. This has been referred to as the \textit{micro-macro loop}, i.e. a bilateral relationship between an emerged symbol system at the macro level and a physical system consisting of communicating and collaborating agents at the micro level \cite{Taniguchi2016}. 
%Therefore, solving the SGP, both in its physical and social aspects, requires an hybrid architecture integrating both approaches, where symbolic representations are learned bottom-up from low-level sensorimotor interaction with the physical and social environment, and are in turn recruited in higher-level top-down control. The reason is that whereas a purely bottom-up approach could explain how symbols are acquired and grounded in the physical interaction with the (social) environment, the function of symbols is to take advantage of abstraction, reasoning and communication abilities in top-down control of the agent behavior. 

Several contributions in social robotics rely on such hybrid architectures integrating bottom-up and top-down processes (e.g.~\cite{scassellati2003investigating, stoytchev2001combining, malfaz2011biologically, scheutzetal13irs}). In \cite{scassellati2003investigating}, an architecture called embodied theory of mind was developed to link high-level cognitive skills to the low-level perceptual abilities of a humanoid and implementing joint attention and intentional state understanding. In \cite{stoytchev2001combining}, or \cite{malfaz2011biologically}, the architecture combines deliberative planning, reactive control, and motivational drives for controlling robots in interaction with humans.

In this paper, we adopt the principles of the \emph{Distributed Adaptive Control} theory of the mind and the brain (DAC, \cite{verschure2003environmentally,verschure2014why}). DAC is a hybrid architecture which posits that cognition is based on the interaction of four interconnected control loops operating at different levels of abstraction (see Figure \ref{fig:wrdac}). The first level is called the \emph{somatic layer} and corresponds to the embodiment of the agent within its environment, with its sensors and actuators as well as the physiological needs (e.g. for exploration or safety). %A specificity of DAC is that the \emph{somatic layer} also includes the physiological needs of the agent (e.g. for exploration or safety) and which drives the dynamics of the whole architecture \cite{puigbo2016towards}.
Extending bottom-up approaches with drive reduction mechanisms, complex behavior is bootstrapped in DAC from the self-regulation of an agent's physiological needs when combined with reactive behaviors (the \emph{reactive layer}). This reactive interaction with the environment drives the dynamics of the whole architecture \cite{puigbo2016towards}, bootstrapping learning processes for solving the physical SGP (the \emph{adaptive layer}) and the acquisition of higher-level cognitive representations such as abstract goal selection, memory and planning (the \emph{contextual layer}). These high-level representations in turn modulate the activity at the lower levels via top-down pathways shaped by behavioral feedback. The control flow in DAC is therefore distributed, both from bottom-up and top-down interactions between layers, as well as from lateral information processing within each layer. 

\subsection{Representation learning for solving the SGP}
\label{sec:solving-sgp}
As we have seen, a cognitive architecture solving the SGP needs to bridge the gap between low-level sensorimotor data and symbolic knowledge. %Taking advantage of such architectures, several interaction paradigms have been proposed for grounding a lexicon in the physical interaction of a robot with its environment. Since the pioneering paradigm of "language games" proposed in \cite{steels1997synthetic}, a number of multi-agent models have been proposed showing how particular properties of language can self-organize out of repeated dyadic interactions between agents of a population (e.g. \cite{Kaplan2000,moulinfrier2015cosmo}). In the domain of HRI, contributions have focused on lexicon acquisition through the transfer of sensorimotor and linguistic information from the interaction between a teacher and a learner through imitation \cite{billard1998grounding}, action \cite{cangelosi2006language,marocco2010grounding} or active exploration \cite{ivaldi2013object}. In all these contributions, solving the SGP requires integrating multimodal information about external world entities (\textit{physical SGP}) with linguistic labels acquired or negotiated through the interaction with social peers (\textit{social SGP}). 
Several methods have been proposed for compressing multimodal signals into symbols. A solution based on geometrical structures was offered by G\"{a}rdenfors with the notion of \textit{conceptual spaces} (e.g., \cite{Gardenfors2000}), whereby similarity between concepts is derived from distances in this space. Lieto et al. \cite{Lieto2017} advocate the use of the conceptual spaces as the \textit{lingua franca} for different levels of representation. 

Another approach has been proposed in \cite{eliasmith2012large,blouw2016concepts}, which involves a single class of mental representations called ``Semantic Pointers''. These representations are particularly suited in solving the SGP as they support binding operations of various modalities, which in turn result in a single representation. This representation (which might have been initially formed by an input of a single modality) can then trigger a corresponding concept, whose occurrence leads to simulated stimuli in the other modalities. Furthermore, while semantic pointers can be represented as vectors, the vector representation can be transformed in neural activity which makes the implementation biologically plausible and allows mapping to different brain areas.

Other approaches consider symbols as fundamentally sensorimotor units. For example, Object-Action Complexes (OACs) build symbolic representations of sensorimotor experience and behaviors through the learning of object affordances \cite{Kruger2011} (for a review of affordance-based approaches, see \cite{Jamone2016}). In \cite{Vernon2015}, a framework founded on joint perceptuo-motor representations is proposed, integrating declarative episodic and procedural memory systems for combining experiential knowledge with skillful know-how. 

%In the proposed \emph{DAC-h3} architecture, there is no fixed interaction paradigm. Instead, the robot and the human are engaged in a mixed-initiative scenario where both can decide to trigger interaction at any time. The robot does it by self-regulating internal drives for social interaction and knowledge acquisition through behavior. 
In DAC-h3, visual, tactile, motor and linguistic information about the present entities is collected proactively through reactive control loops triggering sensorimotor exploration in interaction with a human partner. Abstract representations are learned on-line using state-of-the-art machine learning methods in each modality (see Section~\ref{sec:wrdac}). An entity is therefore represented internally in the robot's memory as the association between abstracted multimodal representations and linguistic labels.

\subsection{Interaction paradigms and autonomous exploration}
\label{sec:autonomous-exploration}
Learning symbolic representations from sensorimotor signals requires an autonomous interaction of a robot with the physical and social world. Several interaction paradigms have been proposed for grounding a lexicon in the physical interaction of a robot with its environment. Since the pioneering paradigm of \textit{language games} proposed in \cite{steels1997synthetic}, a number of multi-agent models have been proposed showing how particular properties of language can self-organize out of repeated dyadic interactions between agents of a population (e.g. \cite{Kaplan2000,moulinfrier2015cosmo}).

In the domain of HRI, significant progress has been made in allowing robots to interact with humans, for example in learning shared plans \cite{lallee2013cooperative, lallee2012towards,petit2013coordinating}, learning to imitate actions \cite{calinon2010learning, demiris2008robot, schaal1999imitation,Lee2013}, and learning motor skills \cite{ewerton2015learning} that can be used for engaging in joint activities. Other contributions have focused on lexicon acquisition through the transfer of sensorimotor and linguistic information from the interaction between a teacher and a learner through imitation \cite{billard1998grounding} or action \cite{cangelosi2006language,marocco2010grounding}. However, in most of these interactions, the human is in charge and the robot is following the human’s lead: the choice of which concept to learn is left to the human and the robot must identify it. In this case, the robot must solve the referential indeterminacy problem described by Quine~\cite{quine1960word}, where the robot language learner has to extract the external concept that was referred to by the human speaker. However, acquiring symbols by interacting with other agents is not only a unidirectional process of information transfer between a teacher and learner \cite{levinson2006human}.

Autonomous exploration and proactive behavior solves this problem by allowing robots to take the initiative in exploring their environment \cite{oudeyer2007intrinsic} and interacting with people \cite{breazeal2000infant}. The benefit of these abilities for knowledge acquisition has been demonstrated in several developmental robotics experiments. In \cite{ivaldi2013object}, it is shown how a combination of social guidance and intrinsic motivation improve the learning of object visual categories in HRI. A similar mechanism is adopted in \cite{Moulin-Frier_Frontiers_2013} for learning complex sensorimotor mappings in the context of vocal development. In \cite{Schrempf2005}, planning conflicts due to the uncertainty of the detected human's intention are resolved by proactive execution of the corresponding task that optimally reduces the system's uncertainty. In \cite{Lutkebohle2009}, the task is to acquire human-understandable labels for novel objects and learning how to manipulate them. This is realized through a mixed-initiative interaction scenario and it is shown that proactivity improves the predictability and success of human-robot interaction.

%In all these contributions, solving the SGP requires integrating multimodal information about external world entities (\textit{physical SGP}) with linguistic labels acquired or negotiated through the interaction with social peers (\textit{social SGP})

A central aspect of the \emph{DAC-h3} architecture is the robot's ability to act proactively in a mixed-initiative scenario. This allows self-monitoring of the robot's own knowledge acquisition process, removing dependence on the human's initiative. Interestingly, proactivity in a sense reverses the referential indeterminacy problem mentioned above by shifting the responsibility of solving ambiguities to the agent who is endowed with the adequate prior knowledge to solve it, i.e., the human, in a HRI context. The robot is now in charge of the concepts it wants to learn, and can use joint attention behaviors to guide the human toward the knowledge it wants to acquire. In the proposed system, this is realized through a set of behavioral control loops, by self-regulating knowledge acquisition, and by proactively requesting missing information about entities from the human partner.

%\cite{forestier2016overlapping,ribes2016active}

\subsection{Language learning, autobiographical memory and narrative expression}
\label{sec:linguisitic-comm}

We have just described the main components that a robot requires to solve the SGP: a cognitive architecture able to process both low-level sensorimotor data and high-level symbolic representation, mechanisms for linking these two levels in abstract multimodal representations, as well autonomous behaviors for proactively interacting with the environment. The final challenge to address concerns the use of the acquired symbols for higher-level cognition, including language learning, autobiographical memory and narrative expression.

Several works address the ability of language learning in robotics. The cognitive architecture of iTalk~\cite{Broz2014} focuses on modeling the emergence of language by learning about the robot's embodiment, learning from others, as well as learning linguistic capability. Cangelosi et al.~\cite{Cangelosi2010} propose that action, interaction and language should be considered together as they develop in parallel, and one influences the others. Antunes et al.~\cite{antunes2016human} assume that language is already learned, and address the issue that linguistic input typically does not have a one-to-one mapping to actions. They propose to perform reasoning and planning on three different layers (low-level robot perception and action execution, mid-level goal formulation and plan execution, and high-level semantic memory) to interpret the human instructions. Similarly,~\cite{krause2014learning} proposes a system to recognize novel objects using language capabilities in one shot. In these works, language is typically used to understand the human and perform actions, but not necessarily to talk about past events that the robot has experienced.

A number of works investigate the expression of past events by developing narratives based on acquired autobiographical %long term memories
memories~\cite{dias2007, Syrdal2014, Sieber2010}. In \cite{Syrdal2014}, a user study is presented which suggests that a robot's narrative allows humans to get an insight into long term human-robot interaction from the robot's perspective. 
The method in~\cite{Sieber2010} takes user preferences into account when referring to past interactions. Similarly to our framework, it is based on the implementation and cooperation between both episodic and semantic memories with a dialog system. However, no learning capabilities (neither language nor knowledge) are introduced by the authors. 

In the proposed \emph{DAC-h3} architecture, the acquired lexicon allows the robot to execute action plans for achieving goal-oriented behavior from human speech requests. Relevant information throughout the interaction of the robot with humans is continuously stored in an autobiographical memory used for the generation of a narrative self, i.e., a verbal description of the own robot's history over the long term (able to store and verbally describe interactions from a long time ago, e.g. a period of several months).%Therefore, we bridge the gap between language acquisition, the formation of an autobiographical memory, and narrative structures.

In the next section, we describe how the above features are implemented in a coherent cognitive architecture made up of functional YARP~\cite{fitzpatrick2008towards} modules running in real-time on the iCub robot.

%% file: 3_cog_arch.tex
\section{The DAC-h3 Cognitive Architecture}
\label{sec:wrdac}

%\tf{General remark: Sometimes the layers start with Capital letters, sometimes with lower case. Personally, I'd suggest to always use lower case letters and italics, e.g. \textit{soma layer}.}

%\cmf{I've entirely revised this section, restructuring it layer by layer and attempting to build a coherent story from the huge set of interacting modules we have. I've kept the existing text as much as possible. Please provide your feedback on it (as constructive as possible and providing ready-to-use pieces of text that solve the mentioned issues whenever possible). I've put comments all over the section pointing at missing information (not that much is missing actually). Everybody is welcome to fill the missing part (some are specific to a specific group though, e.g. action recognition by Sheffield. }

This section presents the \emph{DAC-h3} architecture in detail, which is an instantiation of the DAC architecture for human-robot interaction. The proposed architecture provides a general framework for designing autonomous robots which act proactively for 1) maintaining social interaction with humans, 2) bootstrapping the association of multimodal knowledge with its environment that further enrich the interaction through goal-oriented action plans, and 3) express a verbal narrative. It allows a principled organization of various \emph{functional modules} into a biologically grounded cognitive architecture.

\subsection{Layer and module overview}

In \emph{DAC-h3}, the \emph{somatic layer} consists of an iCub humanoid robot equipped with advanced motor and sensory abilities for interacting with humans and objects. The \emph{reactive layer} ensures the autonomy of the robot through drive reduction mechanisms implementing proactive behaviors for acquiring and expressing knowledge about the current scene. This allows the bootstrapping of adaptive learning of multimodal representations about entities in the \emph{adaptive layer}. More specifically, the \emph{adaptive layer} learns high-level multimodal representations (visual, tactile, motor and linguistic) for the categorization of entities (objects, agents, actions and body parts) and associates them in unified representations. Those representations form the basis of an episodic memory for goal-oriented behavior through planning in the \emph{contextual layer}, which deals with goal representation and action planning. Within the contextual layer, an autobiographical memory of the robot is formed that can be expressed in the form of a verbal narrative.

\begin{figure}[!t]
  \centering
  \includegraphics[width=0.86\linewidth]{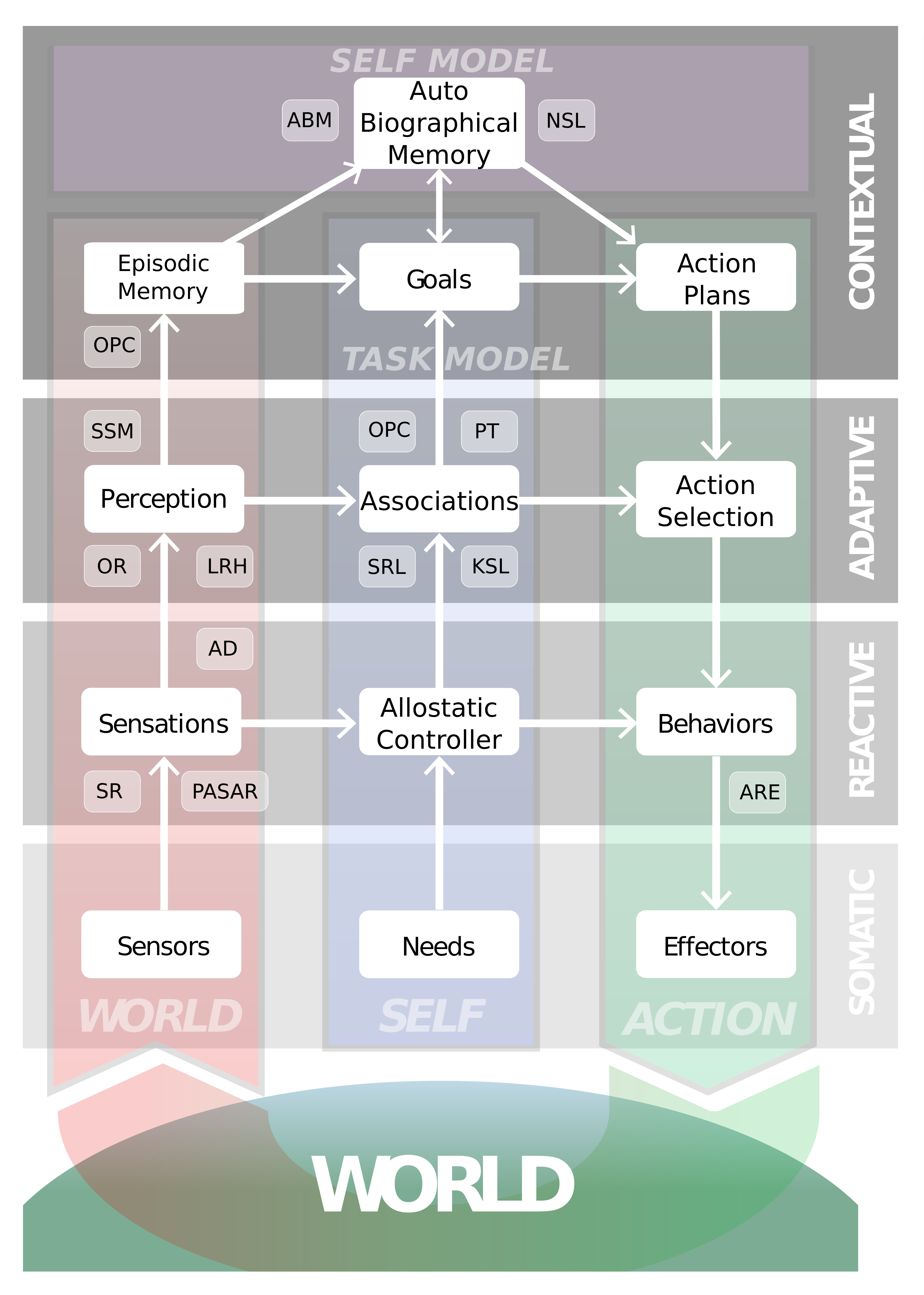}
  %\vspace{-0.2cm}
  \caption{The \emph{DAC-h3} cognitive architecture (see Section \ref{sec:wrdac}) is an implementation of the DAC theory of the brain and mind (see Section~\ref{sec:cogarch-sgp}) adapted for HRI applications. The architecture is organized as a layered control structure with tight coupling within and between layers: the somatic, reactive, adaptive and contextual layers. Across these layers, a columnar organization exists that deals with the processing of states of the world or exteroception (left, red), the self or interoception (middle, blue) and action (right, green). The role of each layer and their interaction is described in Section \ref{sec:wrdac}. 
White boxes connected with arrows correspond to \emph{structural modules} implementing the cognitive modules proposed in the DAC theory. Some of these structural modules rely on \emph{functional modules}, indicated by acronyms in the boxes next to the structural modules. Acronyms refer to the following functional modules. SR: Speech Recognizer; PASAR: Prediction, Anticipation, Sensation, Attention and Response; AD: Agent Detector; ARE: Action Rendering Engine; OR: Object Recognition; LRH: Language Reservoir Handler; SSM: Synthetic Sensory Memory; PT: Perspective Taking; SRL: Sensorimotor Representation Learning; KSL: Kinematic Structure Learning; OPC: Object Property Collector; ABM: Autobiographical Memory; NSL: Narrative Structure Learning.
  }
  %\vspace{-0.25cm}
  \label{fig:wrdac}
\end{figure}

The complete \emph{DAC-h3} architecture is shown in Figure~\ref{fig:wrdac}. It is composed of \emph{structural modules} reflecting the cognitive modules proposed by the DAC theory. Each \emph{structural module} might rely on one or more \emph{functional modules} implementing more specific functionalities (e.g. dealing with motor control, object perception, and scene representation). The complete system described in this section, therefore, integrates several state-of-the-art algorithms for cognitive robotics and integrates them into a structured cognitive architecture grounded in the principles of the DAC theory. In the remainder of this section, we describe each \emph{structural module} layer by layer, as well as their interaction with the \emph{functional modules}, and provide references which provide more detail for the individual modules.
%The RCA is thus divided in 3 columns: World, Self and Action, with 4 levels each: Somatic, Reactive, Adaptive and Contextual. We will now explain for each column how is built our DCA.

%%%%%%%%%%%%%% Fig DAC was originally there: moved to cover 

\subsection{Somatic layer}

%\cmf{Is everybody happy with the soma description?}\tf{Slightly changed, quite happy}

The \emph{somatic layer} corresponds to the physical embodiment of the system. We use the iCub robot, an open source humanoid platform developed for research in cognitive robotics \cite{metta2010icub}. The iCub is a $104\,\text{cm}$ tall humanoid robot with 53 degrees of freedom (DOF). 
%It has two dexterous hands with 19 under-actuated joints and 9 DOF each. 
The robot is equipped with cameras in its articulated eyes allowing stereo vision, and tactile sensors in the fingertips, palms of the hand, arms and torso. The iCub is augmented with an external RGB-D camera above the robot head for agent detection and skeleton tracking. Finally, an external microphone and speakers are used for speech recognition and synthesis, respectively. %[S]might be better to use 'production' instead of 'synthesis'

The \emph{somatic layer} also contains the physiological \emph{needs} of the robot that will drive its reactive behaviors, as described in the following section on the \emph{reactive layer}.

%The first level in the hierarchy of the Self relative to the interoception of the robot, will concern the basics needs of the robot. It corresponds to the sensory-motor interface between the somatic layer and the system. At the level of interaction with the world, there are several components.\mpet{We should state what properties/skills need to be here and then introduce the implementation of each. e.g. "In order to react properly in the world, a robot should detect salient objects or agents, as well as act upon them (including movement, speech, etc.")}

%\noindent%
%\begin{minipage}{\linewidth}
%\captionsetup{type=figure}
%\makebox[\linewidth]{%
%\end{minipage}

\subsection{Reactive layer}
\label{sec:reactive}
Following DAC principles, the \emph{reactive layer} oversees the self-regulation of the internal \emph{drives} of a cognitive agent from the interaction of sensorimotor control loops. The drives aim at self-regulating internal state variables (the \emph{needs} of the \emph{somatic layer}) within their respective homeostatic ranges. In biological terms, such an internal state variable could, for example, reflect the current glucose level in an organism, with the associated homeostatic range defining the minimum and maximum values of that level. A drive for eating would then correspond to a self-regulation mechanism where the agent actively searches for food whenever its glucose level is below the homeostatic minimum and stops eating even if food is present whenever it is above the homeostatic maximum. A drive is therefore defined as the real-time control loop triggering appropriate behaviors whenever the associated internal state variable goes out of its homeostatic range, as a way to self-regulate its value in a dynamic and autonomous way.

%It is common for drives to be competing, which makes clear the need for a method to prioritize them: 
%Let us consider a real-life example: assuming smoking indoors is prohibited, a heavy smoker could experience two pertinent drives during winter --- one for maintaining the body's nicotine level and another for maintaining a reasonable body temperature. Depending on how great their need for nicotine is and how low the temperature is, they will choose to either remain indoors or to stand outdoors in the cold to smoke. In the proposed model, a drive is active only when the associated internal variable is outside its homeostatic range. However, 

In the social robotics context that is considered in this paper, the drives of the robot do not reflect biological needs as above but are rather related to knowledge acquisition and expression in social interaction. At the foundation of this developmental bootstrapping process is the intrinsic motivation to interact and communicate. As described by Levinson \cite{levinson2006human} (see Introduction), a part of the \textit{human interaction engine} is a set of capabilities that include the motivation to interact and communicate through universal (language independent) manners; including looking at objects of interest and at the interaction partner, as well as pointing to these objects. These reactive capabilities are built into the \emph{reactive layer} of the architecture forming the core of the \emph{DAC-h3} interaction engine. These interaction primitives allow the \emph{DAC-h3} system and the human to share attention around specific entities (body parts, objects, or agents), and to bootstrap learning mechanisms in the \emph{adaptive layer} that associate visual, tactile, motor and linguistic representations of entities as described in the next section. 

Currently, the architecture implements the following two drives: one for knowledge acquisition and one for knowledge expression. However, \emph{DAC-h3} is designed in a way that facilitates the addition of new drives for further advancements (see Section~\ref{sec:conclusion}). First, a \emph{drive for knowledge acquisition} provides the iCub with an intrinsic motivation to acquire new knowledge about the current scene. The internal variable associated with this drive is modulated by the number of entities in the current scene with missing information (e.g. unknown name, or missing property). The self-regulation of this drive is realized by proactively asking the human to provide missing information about entities, for instance, their name via speech, synchronized with gaze and pointing; or asking the human to touch the robot's skin associated with a specific body part. %\mpet{We should not restrict the knowledge acquisition to just object labeling. It can also acquire knowledge about properties of entities, e.g. the skin sensors corresponding to a specific body part}\cmf{Agree, I've modified it in this direction.}

Second, a \emph{drive for knowledge expression} allows the iCub to proactively express its acquired knowledge by interacting with the human and objects. %There are two general means of knowledge expression. 
The internal variable associated with this drive is modulated by the number of entities in the current scene without missing information. The self-regulation is then realized by triggering actions toward the known entities, synchronized with verbal descriptions of those actions (e.g. pointing towards an object or moving a specific body part, while verbally referring to the considered entity). 

The implementation of these drives is realized through the three structural modules described below, interacting with each other as well as with the surrounding layers: \emph{1) sensations}, \emph{2) allostatic controller}, and \emph{3) behaviors} (see Figure~\ref{fig:wrdac}). 

\subsubsection{Sensations}
The \emph{sensations} module deals with low-level sensing to provide relevant information for meaning extraction in the \emph{adaptive layer}. Specifically, the module detects presence and position of other agents and their body parts (\emph{agent detector} functional module; of interest are the head location for gazing at the partner and the location of the hands to detect pointing actions) based on the input of the RGB-D camera. Similarly, it detects objects based on a texture analysis and extracts their location using the stereo vision capabilities of the iCub~\cite{Fanello2014}. The \emph{prediction, anticipation, sensation, attention and response} functional module (PASAR;~\cite{Mathews2012}) calculates the saliency of agents based on their motion (increased velocity leads to increased saliency), and similarly the saliency for objects is increased if they move or the partner points at them. Finally, the \emph{speech recognition} functional module extracts text from human speech sensed by a microphone using the Microsoft™ Speech API. The functionalities of the \emph{sensations} module can, therefore, be summarized as dimensionality reduction and saliency computation, and the resulting data are used for bootstrapping knowledge in higher layers of the architecture. %With a given grammar provided by different modules or given by default, this module returns the sentence heard by the robot in textual form.

\subsubsection{Allostatic Controller}
In many situations, several drives which may conflict with each other can be activated at the same time (in the case of this paper, the drive for knowledge acquisition and the drive for knowledge exploration). Such possible conflicts can be solved through the concept of an \emph{allostatic controller} \cite{sanchez2010allostatic,fibla2010allostatic2}, defined as a set of simple homeostatic control loops and dealing with their scheduling to ensure an efficient global regulation of the internal state variables. The scheduling is decided according to the internal state of the robot and the output of the \emph{sensations} module. %, the \emph{allostatic controller} updates the drive levels in real-time and has the role of deciding which drive to regulate at the current time. 
%For example, the knowledge acquisition drive is modulated by the amount of unknown information about the entities present in the current scene, whereas the knowledge expression drive is modulated by the amount of already acquired information about the entities present in the current scene. 
The decision of which drive to follow depends on several factors: the distance of each drive level to their homeostatic boundaries, as well as predefined drive priorities (in \emph{DAC-h3}, knowledge acquisition has priority over knowledge expression, which results in a curious personality).

\subsubsection{Behaviors}
To regulate the aforementioned drives, the \emph{allostatic controller} is connected to the \emph{behaviors} module, and each drive is linked to corresponding behaviors which are supposed to bring it back into its homeostatic range whenever needed. The positive influence of such a drive regulation mechanism on the acceptance of the HRI by naive users has been demonstrated in previous papers \cite{lallee2015grounding,vouloutsi2014influence}.

The drive for knowledge acquisition is regulated by requiring information about entities through coordinated behaviors. Those behaviors depend on the type of the considered entity:
\begin{itemize}
\item In the case of an object, the robot produces speech (e.g. ``What is this object?'') while pointing and gazing at the unknown object.
\item In the case of an agent, the robot produces speech (e.g. ``Who are you?'') while looking at the unknown human.
\item In the case of a body part, the robot either asks for the name (e.g. ``How do you call this part of my body?'') while moving it or, if the name is already known from a previous interaction, asks the human to touch the body part while moving it (e.g., ``Can you touch my index while I move it, please?'').
\end{itemize}
The multimodal information collected through these behaviors will be used to form unified representations of entities in the \emph{adaptive layer} (see next section).

The drive for knowledge expression is regulated by executing actions towards known entities, synchronized with speech sentences parameterized by the entities' linguistic labels acquired in the \emph{adaptive layer} (see next section). %\cmf{Do we have more to add here? Do we e.g. express knowledge about body parts or agents?}\mpet{We have, but not yet trigger by the knowledge expression per se I think (which just randomly pick an object to express knowledge, nothing else). The agent name is used for all the future interaction and the bodypart name + babbling is used for joint attention when investigating the touch property}
Motor actions are realized through the \emph{action rendering engine} (ARE \cite{PattaciniIROS2010}) functional module which allows executing complex actions such as push, reach, take, look in a coordinated human-like fashion. % in terms of primitive motor movements in both the joint and the Cartesian space. The ARE system takes as input the label of the requested action (e.g. push) along with the 3D coordinates of the object (provided by the \emph{sensations} module) over which the action needs to be executed. 
%The whole trajectory gets decomposed in multiple trajectories via points whose intermediate movements are resolved by means of a non-linear constraints optimization, and then performed by a multi-referential operational controller as described in \cite{PattaciniIROS2010}. The robot executes the motor actions moving its limbs in a coordinated human-like fashion. 
Language production abilities are implemented in the form of predefined grammars (for example asking for the name of an object). Semantic words associated to entities are not present at the reactive level, but are provided from the learned association operating in the \emph{adaptive layer}. 
The \emph{iSpeak} module implements a bridge between the iCub and a voice synthesizer %(e.g. Festival\footnote{\url{http://www.cstr.ed.ac.uk/projects/festival/}}, Acapela\footnote{\url{http://www.acapela-group.com/}}) 
%\tf{Refs added}
by synchronizing the produced utterance with the lip movements of the iCub to realize a more vivid interaction~\cite{Parmiggiani2012}.

%\mpet{However what exactly is tagging? the tagging behaviour is actually quite complex and seems to be more in the contextual part: it is an action plans links together with dialogue. e.g. tagging joint 9: 1. I move the joint 9 2. I ask the human the name 3. I update my knowledge}\cmf{The learning aspect of tagging is now described in the adaptive layer. It's sequential aspect is described in the contextual section.}\tf{Can we remove these comments?}

%In this paper, the focus is on its integration with an \emph{adaptive} and \emph{contextual layer}, as described in the next two sections, providing a complete cognitive architecture for HRI.

\subsection{Adaptive layer}
\label{sec:adaptive}

The \emph{adaptive layer} oversees the acquisition of a state space of the agent-environment interaction by binding visual, tactile, motor and linguistic representations of entities. It integrates functional modules for maintaining an internal representation of the current scene, visually categorizing entities, recognizing and sensing body parts, extracting linguistic labels from human speech, and learning associations between multimodal representations. They are grouped in three structural modules described below: \emph{perceptions}, \emph{associations} and \emph{action selection} (see Figure~\ref{fig:wrdac}).

%The adaptive layer is in charge of acquiring a state space of the agent-environment interaction through learning mechanisms in order to solve the so-called symbol grounding problem~\cite{Harnad1990}. In the proposed architecture, it consists in learning visual representations of entities (objects, actions and agents) detected in the reactive layer and to associate them with their linguistic labels provided by the human. The adaptive layer also deals with learning body representations, associating them with motor and touch events as well as their linguistic labels. That layer is finally responsible of transforming meaning representations to speech sentences in a bidirectional way. \cmf{Maybe language should be at the contextual level actually, to see.}

\subsubsection{Perceptions}
\label{sec:perceptions}
The \emph{object recognition} functional module~\cite{pasquale2016} 
%\tf{I renamed IOL to object recognition module, as we don't use anything but the object recognition from IOL.} 
is used to learn the categorization of objects directly from the visual information given by the iCub eyes with resort to the most recent deep convolutional networks. 
%It provides a complete chain to address the real-world object recognition problem by coding the images with resort to the most recent deep convolutional networks, and successively applying a support vector machine linear classification. The input to this system are images acquired from the robot cameras containing the objects we aim to recognize. Objects are segmented out of the background relying on the local binary pattern technique; 
The bounding boxes of the objects found in the \emph{Sensations} module are fed to the learning module for the recognition stage. The output of the system consists of the 2D (in the image plane) and 3D (in the world frame) positions of the identified objects along with the corresponding classification scores as stored in the \emph{objects properties collector} memory (explained below).

%\cmf{The paragraph below is quite long compared to others, it might be useful to remove unnecessary information}
There are two functional modules related to language understanding and language production, both integrated within the \emph{language reservoir handler} (LRH). The \emph{comprehension of narrative discourse module} receives a sentence and produces the representation of the corresponding meaning, and can thus transform human speech into meaning. The \emph{module for narrative discourse production} receives a representation of meaning and generates the corresponding sentence (meaning to speech). The meaning is represented in terms of PAOR: \emph{predicate(arguments)}, where \emph{arguments} correspond to thematic roles \emph{(agent,object,recipient)}. Both models are implemented as recurrent neuronal networks based on reservoir computing \cite{hinaut2013real, hinaut2015exploring, mealier2016}.

The \emph{synthetic sensory memory} (SSM) module is currently employed for face recognition and action recognition using a fusion of RGB-D data and object location data as provided from the \emph{sensations} module. In terms of action recognition, SSM has been trained to automatically segment and recognize the following actions: push, pull, lift, drop, wave, and point, while also actively recognizing if the current action is known or unknown. 
More generally, it has been shown that SSM provides abilities for pattern learning, recall, pattern completion, imagination and association~\cite{damianou2015top}. The SSM module is inspired by the role of the hippocampus by fusing multiple sensory input streams and representing them in a latent feature space \cite{damianou2013deep}. %, which emerges as \emph{Deep Gaussian Processes}  are used as the underlying technology. 
During recall, SSM performs classification of incoming sensory data and returns a label along with an uncertainty measure corresponding to the returned label, which is a use case of the action and face recognition tasks within \emph{DAC-h3}. SSM is also capable of imagining novel inputs or reconstructing previously encountered inputs and sending the corresponding generated sensory data. This allows for the replay of memories as detailed in \cite{camilleri2016icub}. 
%In \emph{DAC-h3}, SSM is currently employed for face recognition and action recognition using a fusion of RGB-D data and object location data. 

\subsubsection{Associations}

The \emph{associations} structural module produces unified representations of entities by associating the multimodal categories formed in the \emph{perception} module. Those unified representations are formed in the \emph{objects properties collector} (OPC), a functional module storing all information associated with a particular entity at the present moment in a proto-language format as detailed in \cite{lallee2015grounding}. An entity is defined as a concept which can be manipulated and is thus the basis for emerging knowledge. In \emph{DAC-h3}, each entity has a name associated, which might be unknown if the entity has been discovered but not yet explored. More specifically, higher level entities such as objects, body parts and agents have additional intrinsic properties. For example, an object also has a location and dimensions associated with it. Furthermore, whether the object is currently present is encoded as well, and if so, its saliency value (as computed by the PASAR module described in Section~\ref{sec:reactive}). On the other hand, a body part is an entity which contains a proprioceptive property (i.e. a specific joint), and a tactile information property (i.e. association with tactile sensors). Thus, the OPC allows integrating multiple modalities of one and the same entity to ground the knowledge about the self, other agents, and objects, as well as their relations. Relations can be used to link several instances in an ontological model (see Section~\ref{sssec:EM}: \textit{Episodic Memory}).
%\cmf{OPC to be described here%, with an emphasis on how it allows the integration of multimodal representations of entities. See also "Episodic Memory" in contextual layer
%} \tf{Done}

Learning the multimodal associations that form the internal representations of entities relies on the behavior generated by the knowledge acquisition drive operating at the \emph{reactive level} (see previous section). Multimodal information about entities generated by those behaviors is bound together by registering the acquired information in the specific data format used by the OPC. For instance, the \textit{language reservoir handler} module described above deals with speech analysis to extract entity labels from human replies (e.g. ``this is a cube''; \{P:is, A:this, O:cube, R:$\emptyset$\}). The extracted labels are associated with the acquired multimodal information which depends on the entity type: visual representations generated by the \emph{object recognition} module in case of an object or \textit{agent detector} in case of an agent, as well as motor and touch information in case of a body part. 

The associations of representations can also be applied to the developmental robot itself (instead of external entities as above), to acquire motor capabilities or to learn the links between motor joints and skin sensors of its body~\cite{ZambelliIROS2016W}.
Learning self-related representations of the robot's own body schema is realized by the \emph{sensorimotor representation learning} functional module dedicated to forward model learning \cite{Zambelli2016tcds}. The module receives sensory data collected from the robot’s sensors (e.g. cameras, skin, joint encoders) and allows accurate prediction of the next state given the current state and an action. Importantly, the forward model is learned based on sensory experiences rather than based on known mechanical properties of the robot's body. %by means of  an online heterogeneous ensemble of predictors \cite{Zambelli2016tcds}. This learning method achieves predictions which are more accurate compared with single models' alternatives by combining multiple predictors of different types. The ensemble includes echo-state networks \cite{jaeger2002adaptive}, online echo state Gaussian processes \cite{soh2015spatio} and locally weighted projection regression models \cite{vijayakumar2000fast}. 
%an on-line echo state Gaussian process algorithm \cmf{Any reference on this?}\tf{@Martina: Please put your TCDS reference here}. 

%Sensory predictions are then produced using the self-learned forward model implemented by the proposed ensemble method.
%\cmf{OESGP to be defined}\tf{@Martina: Instead of OESGP, I'd describe your ensemble here. Can you please do it?}
%\mh{Even if the module is called ``body schema'' it may be confusing to the reader to call it that way. What role is it exactly playing in the demo?}\mpet{During the demo, the first module "babbling" (BS-B) is called to produce babbling fingers allowing the human to label the body part. Also, maybe BS-SP should be in Adaptive-Self as it concern prediction of self-sensors?}\cmf{It is now the case, BS in Adaptive-Self and SP in Adaptive-World. Please check it for consistency.}

\begin{figure}[t]
    \centering
        \includegraphics[width=0.375\textwidth]{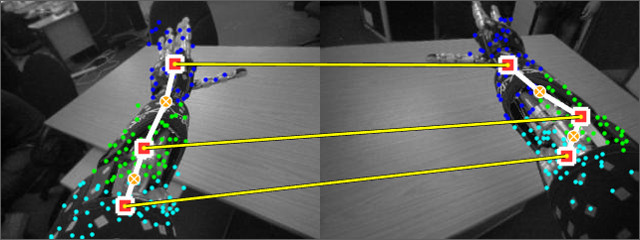}\\
%\vspace{0.1cm}
\includegraphics[width=0.375\textwidth]{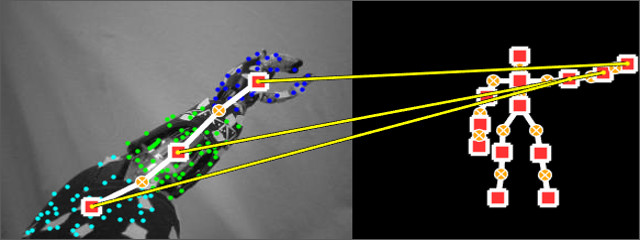}
%\vspace{-0.2cm}
    \caption{
    Examples of the kinematic structure correspondences which the iCub has found. The top figure shows the correspondences between the left and right arm of the iCub, which can be used to infer the body part names of one arm if the corresponding names of the other arm are known. Similarly, the bottom figure shows correspondences between the robot's body and the human's body.
    \label{fig:kinematic-structure}}
\end{figure}

The \emph{kinematic structure learning} functional module~\cite{ChangTPAMIKinematicStructure,ChangCVPR2016} estimates an articulated kinematic structure of arbitrary objects (including the robot's body parts and humans) using visual input videos of the iCub eye cameras. This again is based on sensory experiences rather than known properties of the agents, which is important to autonomously identify the abilities of other agents. %By combining estimated motion and skeleton information, it generates elaborate and plausible kinematic structures. 
% The input is given as a set of sequential images retrieved from the iCub eye cameras, and the output is an estimated structure result that is overlaid on the input video. 
% The kinematic structure estimation is performed by an adaptive motion segmentation through iterative fine-to-coarse merging, skeleton estimation from sparse feature points, generating tree structured kinematic structure using the motion segments and skeleton information.
% \tf{@Hyung Jin this doesn't make much sense yet, please check this.}
Based on the estimated articulated kinematic structures~\cite{ChangTPAMIKinematicStructure}, we also allow the iCub to anchor two objects’ kinematic structure joints by observing their movements~\cite{ChangCVPR2016} and formulating the problem of finding corresponding kinematic joint matches between two articulated kinematic structures. % via hypergraph matching, whilst being accurate and plausible under appearance and motion variations. 
%To the best of our knowledge this is the first work on matching correspondences between dynamic structures. We develop a novel hypergraph matching method which is capable of establishing correspondences between articulated kinematic structures considering newly designed similarity measures. 
%The similarity measures consider structural topology (first order), kinematic correlation (second order) and combinatorial motion (third order) similarities simultaneously. 
This allows the iCub to infer correspondences between its own body parts (its left arm and its right arm), as well as between its own body and the body of the human as retrieved by the \emph{agent detector}~\cite{ZambelliIROS2016W} (see Figure~\ref{fig:kinematic-structure}).
%\tf{If we still have space, we could add one or two figures here.}
%and incorporate them into the hypergraph matching framework with weight normalisation. We introduce a new topologically constrained subgraph isomorphism measure for the structural similarity and a restricted combinatorial motion descriptor for the motion similarity.

Finally, based on these correspondences, the \emph{perspective taking} functional module~\cite{FischerICRA2016} enables the robot to reason about the state of the world from the partner's perspective. This is important in situations where the views of the robot and the human diverge, for example, due to objects which are hidden to the human but visible to the robot. More importantly, perspective taking is thought to be an essential element for successful cooperation and to ease communication, for example by resolving ambiguities~\cite{Johnson2005}. By mentally aligning the self-perspective with that of the human partner, this module allows algorithms (concerned with the visuospatial perception of the world) to reason as if the input was acquired from an egocentric perspective; which allows to use learning algorithms trained on egocentric data to reason on data acquired from the human's perspective without the need of adapting them. %Interestingly, it was proposed that there are two separate processes involved in perspective taking depending on the difficulty of the task~\cite{Michelon2006}. We follow this proposal and solve rather simple tasks such as determining whether an object is visible to the human using line of sight tracing. Complex tasks such as visually imagining the world from the human's point of view are solved by a mental rotation, which allows algorithms to reason as if the input data was acquired from an egocentric perspective. Therefore, learning algorithms which were trained with egocentric data can potentially be applied to reason on data acquired from the other agents perspective without the need of adaptation, as long as the data can be transformed using the mental rotation abilities.

%For this, the point cloud acquired from the RGB-D camera is rotated such that the origin coincides with the head of the human. Furthermore, the mental rotation allows algorithms to reason as if the input data was acquired from an egocentric perspective, which can be used for left/right judgments of object locations. As input, the system receives the object locations; the estimated head pose of the partner; and the point cloud acquired from the RGB-D camera. The system produces an output that characterizes the visibility of each object, the spatial location of each object (left / right of human), as well as a reconstructed view from the human's perspective. This can then be used when reasoning about the object which is referred to by the human. 

\subsubsection{Action Selection}
The \emph{action selection} module uses the information from \emph{associations} to provide context to the \emph{behaviors} module at the reactive level. This context corresponds to entity names which are provided as parameters to the \emph{behaviors} module, for instance pointing at a specific object or using the object linguistic label in the parameterized grammars defined at the reactive level. This module also deals with the scheduling of action plans from the \emph{contextual layer} according to the current state of the system as explained in the following.

\subsection{Contextual layer}
\label{sec:contextual}

The \emph{contextual layer} deals with higher-level cognitive functions that extend the time horizon of the cognitive agent, such as an episodic memory, goal representation, planning and the formation of a persistent autobiographical memory of the robot interaction with the environment. These functions rely on the unified representations of entities acquired at the \emph{adaptive level}. The \emph{contextual layer} consists of three functional modules that are described below: \emph{1) episodic memory}, \emph{2) goals and action plans}, and \emph{3) autobiographical memory} used to generate a narrative structure. 

\subsubsection{Episodic Memory}\label{sssec:EM}
%\cmf{References to be included properly once we are happy with this subsection}
The \emph{episodic memory} relies on advanced functions of the \textit{object property collector} (OPC) to store and associate information about entities in a uniform format based on the interrogative words ``who'' (is acting), ``what'' (they are doing), ``where'' (it happens), ``when'' (it happens), ``why'' (it is happening) and ``how'' called an H5W data structure~\cite{lallee2015grounding}. It is used for goal representation and as elements of the \emph{autobiographical memory}. 
%Formalizing the content and evolution of a scene requires the combination of perceptual, symbolic and rule based reasoning in a single unified framework. Such processes will generate information about who is acting, what they are doing, where and when it happens, and this will give cues about why it is happening. 
H5W have been argued to be the main questions any conscious being must answer to survive in the world \cite{Prescott2014,Verschure2013}.

The concept of \emph{relations} is the core of the H5W framework. It links up to five concepts and assigns them with semantic roles to form a solution to the H5W problem. We define a \emph{relation} as a set of five edges connecting those nodes in a directed and labeled manner. The labels of those edges are chosen so that the \emph{relation} models a typical sentence from the English grammar of the form: Relation $\rightarrow$ Subject Verb [Object] [Place] [Time]. The brackets indicate that the components are optional; the minimal relation is therefore composed of two entities representing a subject and a verb. 
%\cmf{Related aspects of SAM could fit here as well}. 

\subsubsection{Goals and action plans}
Goals can be provided to the iCub from human speech, and a meaning is extracted by the \textit{language reservoir handler}, forming the representation of a goal in the \textit{goals} module. Each goal consequently refers to the appropriately predefined action plan, defined as a state transition graph with states represented by nodes and actions represented by edges of the graph. The \textit{action plans} module extracts sequences of actions from this graph, with each action being associated with a pre- and a post-condition state. Goals and action plans can be parameterized by the name of a considered entity. For example, if the human asks the iCub to take the cube, this loads an action plan for the goal ``Take an object'' which consists of two actions: ``Ask the human to bring the object closer'' and ``Pull the object''. In this case, each action is associated with a pre- and post-condition state in the form of a region in the space where the object is located. In the \textit{action selection} module of the adaptive layer, the plan is instantiated toward a specific object according to the knowledge retrieved from the \textit{associations} module (e.g. allowing to retrieve the current position of the cube). The minimal sequence of actions achieving the goal is then executed according to the perceived current state updated in real-time, repeating each action until its post-condition is met (or cease making the effort after a predefined timeout).

Although quite rigid in its current implementation, in the sense that action plans are predefined instead of being learned from the interaction, this planning ability allows closing the loop of the whole architecture, where drive regulation mechanisms at the \emph{reactive layer} can now be bypassed through contextual goal-oriented behavior. Limitations of this system are discussed in Section~\ref{sec:conclusion}: Conclusions.

\subsubsection{Autobiographical Memory}
%\cmf{proofreading this section would be useful}\tf{@Maxime and I did some changes, we think it's better now.}
%The autobiographical memory collects long term information (days, months, years) about interactions by taking snapshots of the environmental information from the \emph{Episodic Memory}, and generating high level knowledge obtained through reasoning in a Synthetic Semantic Memory (SSM). Memory information is stored in a SQL database. The ABM is bio-motivated based on the human Declarative Long Term Memory situated in the medial temporal lobe, and the distinction facts/events \cite{Pointeau2014TAMD,Petit2016TCDS} \tf{Not sure what you mean with "and the distinction facts/events"}.

The \textit{autobiographical memory} (ABM \cite{Pointeau2014TAMD, Petit2016TCDS, PetitFischerIROS2016W}) collects long term information (days, months, years) about interactions motivated by the human declarative long term memory situated in the medial temporal lobe, and the distinction between facts and events \cite{Pointeau2014TAMD}. It stores data (e.g. objects locations, human presence) from the beginning to the end of an episode by taking snapshots of the environmental information from the \emph{episodic memory} containing the pre-conditions and effects of episodes. This allows the generation of high-level concepts extracted by knowledge-based reasoning. 
%The memory information itself is stored in a SQL database. 
In addition, the ABM captures continuous information during an episode (e.g. images from the camera, joints values)~\cite{Petit2016TCDS}, which can be used by reasoning modules focusing on the action itself, leading to the production of a procedural memory (e.g. through learning from motor babbling or imitation)~\cite{PetitFischerIROS2016W}.

%The complimentary \emph{ABM reasoning module} analyses and clusters the information contained in the \emph{episodic memory} and creates the corresponding knowledge in the SSM. These reasoning functions can be applied on spatial, temporal, contextual or conceptual information \cite{Pointeau2014TAMD,Petit2016TCDS,pointeau2013embodied}. The system can receive queries for knowledge information concerning reasoning (i.e.: ``howTo'', ``whatIs'' ...). Through clustering, reasoning and statistical analysis of the ABM contents, the system generates answers to these queries \tf{@Greg: Are you using this reasoning module for the narrative part? If not we should remove this}.

The \textit{narrative structure learning module} builds on the language processing and ABM capabilities. Narrative structure learning occurs in three phases: 1) First the iCub acquires experience in a given scenario, which generates the meaning representation in the ABM. 2) The iCub then formats each story in term of initial states, goal states, actions and results (IGARF graph \cite{mealier2016}). 3) The human then provides a narration (that is understood using the reservoir system explained in Section~\ref{sec:adaptive}) for the scenario. 

By mapping the events of the narration to the event of the story, the robot can extract the meaning of different discourse functions words (such as ``because''). It can thus automatically generate the corresponding form-meaning mapping that defines the individual grammatical constructions and their sequencing that defines the narrative construction of a new narrative.

\subsection{Summary on the DAC-h3 architecture}
\label{sec:layer_synergy}
The DAC-h3 architecture, therefore, integrates several state-of-the-art algorithms for cognitive robotics and integrates them into a structured cognitive architecture grounded in the principles of the DAC theory. The drive reduction mechanisms in the \emph{reactive layer} allow a complex control of the iCub robot which proactively interacts with humans. In turn, this allows the bootstrapping of adaptive learning of multimodal representations about entities in the \emph{adaptive layer}. Those representations form the basis of an episodic memory for goal-oriented behavior through planning in the \emph{contextual layer}. The life-long interaction of the robot with humans continuously feed an autobiographical memory able to retrieve past experience from request and to express it verbally in a narrative. Altogether, this allows the iCub to interact with humans in complex scenarios, as described in the next section.

%\cmf{Please check that the following paragraph is still coherent (the part on motor babbling) with the new Section IV. If not we can consider either removing it, or adapting section IV accordingly}\tf{I'll move it.}
%This is of particular interest for the robot to show a proactive behavior when delayed reasoning is involved. Consider the scenario where a computationally expensive algorithm is employed which cannot run in real-time. The robot can decouple the data acquisition from the reasoning process (both can be done without human supervision), as well as from the semantic labeling (which requires the presence of the human teacher). One example in the next section is motor babbling for data acquisition, the extraction of the kinematic structure of the iCub hand with the kinematic structure learning functional module, and the decoupled labeling of body parts based on the kinematic structure which is provided during the human-robot interaction.
%\mpet{Newly added.}

%if acquire new knowledge then this new stuff can be used to reason about former memory + example

%% file: 4_experiments.tex
\section{Experimental Results}
\label{sec:results}

This section validates the cognitive architecture described in the previous section on a real demonstration with an iCub humanoid robot interacting with objects and a human. We first describe the experimental setup, followed by the behaviors provided to the robot (self-generated and human-requested). Finally, we analyze the DAC-h3 system in two ways: a complete version reporting the full complexity of the system through multiple video demonstrations and the detailed analysis of a particular interaction, as well as a simplified version showing the effect of the robot's proactivity level on naive users.

The code for reproducing these experiments on any iCub robot is available open-source at \url{https://github.com/robotology/wysiwyd}. It consists of all modules described in the last section implemented in either C++ or Python, and relies on the YARP middleware \cite{fitzpatrick2008towards} for defining their connections and ensuring their parallel execution in real-time.

\begin{figure}
    \centering
        \includegraphics[width=0.4\textwidth]{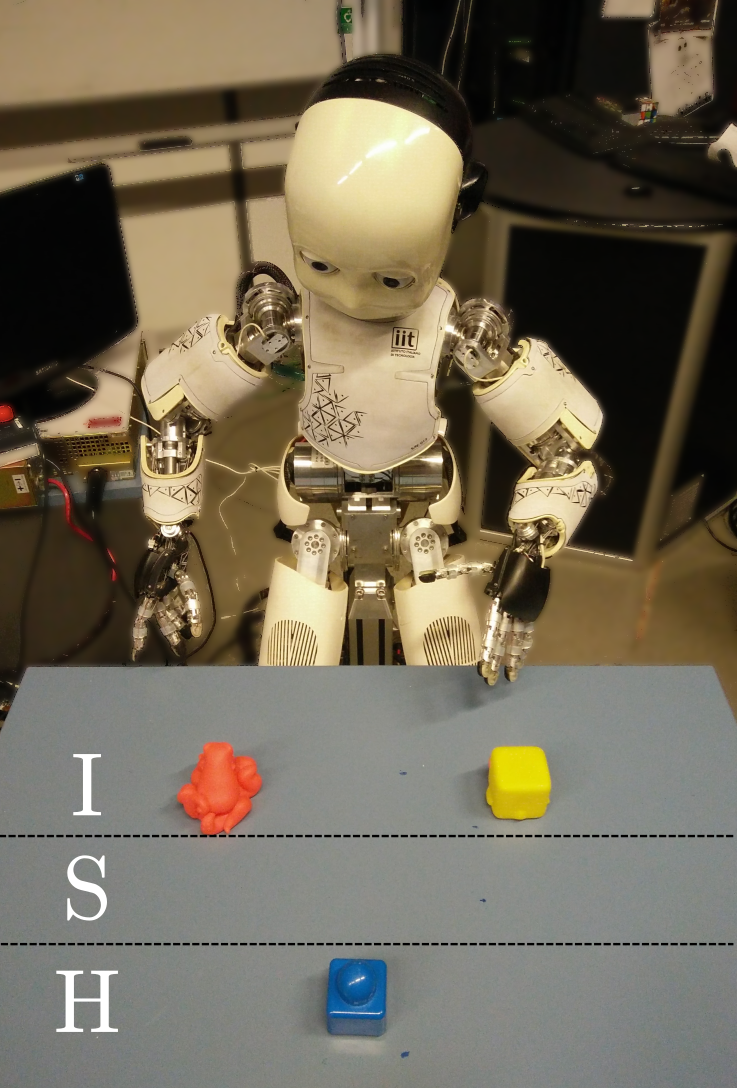}
    \caption{The setup consists of an iCub robot interacting with objects on the table and a human in front of it. The table is separated (indicated by horizontal lines) into three areas: $I$ for the area only reachable by the iCub, $S$ for the shared area, and $H$ for the human-only area (compare with Section~\ref{expsetup}). \label{fig:setup}}
  %  \vspace{-0.2cm}
\end{figure}

\subsection{Experimental setup}
\label{expsetup}
We consider an HRI scenario where the iCub and a human face each other with a table in the middle and objects placed on it. 
The surface of the table is divided into three distinct areas, as shown in Figure~\ref{fig:setup}:
\begin{enumerate}
\item an area which is only reachable by the iCub ($I$),
\item an area which is only reachable by the human ($H$), and
\item an area which is reachable by both agents ($S$ for \textit{Shared}).
\end{enumerate}

The behaviors available to the iCub are the following:

\begin{itemize}
\item ``Acquire missing information about an entity'', which is described in more detail in Section~\ref{sec:knowledge_exploration}.
\item ``Express the acquired knowledge'', which is described in more detail in Section~\ref{sec:knowledge_expression}.
\item ``Move an object on the table'', either by pushing it from region $I$ to $S$ or pulling it from region $S$ to $I$, 
\item ``Ask the human to move an object'', either by asking to push the object from region $H$ to $S$ or by asking to pull it from region $S$ to $H$.
\item ``Show learned representations on screen'' while explaining what is being shown, e.g. displaying the robot kinematic structure learned from a previous arm babbling phase.
\item ``Interact verbally with the human'' while looking at her/him. This is used for replying to some human requests as described in Section~\ref{sec:goalexecution}. 
\end{itemize}

These behaviors are implemented in the \textit{behaviors} module and can be triggered from two distinct pathways as shown in Figure~\ref{fig:wrdac}. The behaviors for acquiring and expressing knowledge are triggered through the drive reduction mechanisms implemented in the \textit{allostatic controller} (Section~\ref{sec:reactive}) and are self-generated by the robot. The remaining behaviors are triggered from the \textit{action selection} module (Section~\ref{sec:adaptive}), scheduling action sequences from the \textit{goals} and \textit{action plans} modules (Section~\ref{sec:contextual}). In the context of the experiments described in this section, these behaviors are requested by the human partner. We describe these two pathways in the two following subsections.

\subsection{Self-generated behavior}
\label{sec:autonomous}

Two drives for knowledge acquisition and knowledge expression implement the interaction engine of the robot (see Section~\ref{sec:reactive}). They regulate the knowledge acquisition process of the iCub and proactively maintain the interaction with the human. The generated sensorimotor data feeds the adaptive layer of the cognitive architecture to acquire multimodal information about the present entities (see Section \ref{sec:adaptive}). In the current experiment, the entities are objects on the table, body parts (fingers of the iCub), human partners, and actions. The acquired multimodal information depends on the considered entity. Object representations are based on visual categorization and stereo-vision based 3D localization performed by the \textit{object recognition} functional module. Body part representations associate motor and touch events. Agents and actions representations are learned from visual input in the \textit{synthetic sensory memory} module presented in Section~\ref{sec:perceptions}. Each entity is also associated with a linguistic label learned by self-regulating the two drives detailed below.

\subsubsection{Drive to acquire knowledge}
\label{sec:knowledge_exploration}
This drive maintains a curiosity-driven exploration of the environment by proactively requesting the human to provide information about the present entities, e.g. naming an object or touching a body part. The drive level decays proportionally to the amount of missing information about the present entities (e.g. the unknown name of an entity). When below a given threshold, it triggers a behavior following a generic pattern of interaction, instantiated according to the nature of the knowledge to be acquired. It begins with a behavior to obtain a joint attention between the human and the robot toward the entity that the robot wants to learn about. After the attention has been attracted toward the desired entity, the iCub asks for the missing information (e.g. the name of an object or of the human, or in the case of a body part the name and touch information) and the human replies accordingly. In a third step, this information is passed to the adaptive layer and the knowledge of the robot is updated in consequence.

Each time the drive level reaches the threshold, an entity is chosen in a pseudo-random way within the set of perceived entities with missing information, with a priority to request the name of a detected unknown human partner. Once a new agent enters the scene, the iCub asks for her/his name, which is stored alongside representations of its face in the \textit{synthetic sensory memory} module. Similarly, the robot stores all objects it has previously encountered in its \textit{episodic memory} implemented by the \textit{object property collector} module. When the chosen entity is an object, the robot asks the human to provide the name of interest while pointing at it. Then, the visual representation of the object computed by the \textit{object recognition} module is mapped to the name. 
When the chosen entity is a body part (left-hand fingers), the iCub first raises its hand and moves a random finger to attract the attention of the human. Then it asks for the name of that body part. This provides a mapping between the robot's joint identifier and the joint's name. 
This mapping can be extended to include tactile information by asking the human to touch the body part which is being moved by the robot.

Once a behavior has been triggered, the drive is reset to its default value and decays again as explained above (the amount of the decay being reduced according to what has been acquired).

\subsubsection{Drive to express knowledge}
\label{sec:knowledge_expression}
This drive regulates how the iCub expresses the acquired knowledge through synchronized speech, pointing and gaze. It aims at maintaining the interaction with the human by proactively informing her/him about its current state of knowledge. The drive level decays proportionally to the amount of already acquired information about the present entities. When below a given threshold (meaning that a significant amount of information has been acquired), it triggers a behavior alternating gazing toward the human and a known entity, synchronized with speech expressing the knowledge verbally, e.g. ``This is the octopus'', or ``I know you, you are Daniel''. Once such a behavior has been triggered, the drive is reset to its default value and decays again as explained above (the amount of the decay changing according to what is learned by satisfying the drive for knowledge acquisition).

These two drives allow the robot to balance knowledge acquisition and expression in an autonomous and dynamic way. At the beginning of the interaction, the robot has little knowledge about the current entities and therefore favors behaviors for knowledge acquisition. By acquiring more and more knowledge, it progressively switches to behaviors for knowledge expression. If new entities are introduced, e.g. a new object or another human, it will switch back to triggering more behaviors for knowledge acquisition and so on.

\subsection{Human-requested behavior}
\label{sec:goalexecution}
The representations which are acquired through satisfying the drives introduced above allow a more complex interaction through goal-oriented behavior managed by the contextual layer (see Figure~\ref{fig:wrdac} and Section~\ref{sec:contextual}). Goals can be provided to the iCub from human speech and a meaning is extracted by the \textit{language reservoir handler}, forming the representation of a goal in the \textit{goals} module. Each goal is associated with an action plan on the form of a sequence of actions together with their pre- and post-conditions in the \textit{action plans} module. The \textit{action selection} module takes care of the execution of the plan according to the \textit{associations} known to the robots, triggering the appropriate behaviors according to its current perception of the scene updated in real time. Goal achievement bypasses the reactive behavior described in the previous subsection by freezing all the drive levels during the execution of the plan. The available goals are described below. 

\subsubsection{Give or take an object}
These goals are generated from a human verbal request, e.g. \textit{``Give me the octopus''} or \textit{``Take the cube''}. Here, the goal is represented as a region on the table, either the human area $H$ (for the ``Give'' goal) or the iCub area $I$ (for the ``Take'' goal), where the mentioned object should be placed. Action plans are generated from the state-transition graph shown in Figure~\ref{fig:state-transitions}. State perception is updated in real-time according to the perceived location of the object computed through stereo-vision in the \textit{object recognition} module. 

\subsubsection{Point to an object}
This goal is generated through a verbal request, e.g. \textit{``Point to the octopus''}. If the mentioned object is not known to the iCub, it will first ask the human to point to it to learn the new association between the name and the object's visual representation. Once the name is known, or if it was already known, the iCub will point to the object. 

\subsubsection{Say the name of a recognized action}
This goal is generated through a verbal request, e.g. \textit{``How do you call this action?''} formulated just after the human has performed an action on an object. Six actions can be recognized by the \textit{synthetic sensory memory} module: ``push'', ``pull'', ``lift'', ``drop'', ``wave'', and ``point''. The reply from the iCub provides the name of the action and the object as well as the hand used by the human, e.g. \textit{``You pushed the cube with your left hand''}.

\begin{figure}[t]
    \centering
        \includegraphics[width=0.3\textwidth]{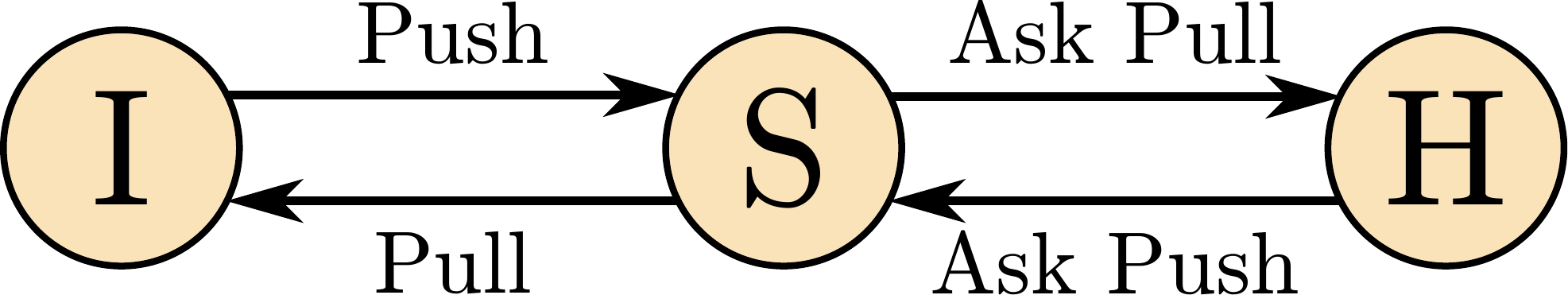}
    \caption{State transition graph used for generating the action plans of the goals ``Give'' and ``Take''. Nodes correspond to the table regions indicated in Figure~\ref{fig:setup}. Arrows correspond to the actions to be executed for realizing a transition. According to the observed current state of the object, the iCub will execute the action which brings it closer to the goal state. For example, if the goal is to take an object which is in the human area, the robot will first ask the human to push it, and subsequently pull it into its own area. \label{fig:state-transitions}}
\end{figure}

\subsubsection{Say what happened during a past interaction with a human}
This goal is generated through a verbal request, e.g. \textit{``What have you done the other day?''}. Based on its past interactions with the environment and with humans, the iCub has stored all the relevant information in its \textit{autobiographical memory} (see Section~\ref{sec:contextual}), including its own drives, motivations, and actions, as well of the actions of the human (both, spoken and physically enacted). The \textit{narrative handler} functional module can generate a narrative discourse from the content of the \textit{autobiographical memory} and to generate an action plan in the form of a sequence of sentences. The human can request more detailed information about an event using sentences like \textit{``What happened next?''} or \textit{``Why did you do that?''}, this later question being answered according to the stored drive levels and goal states of the considered events. The robot can also learn a set of questions that it can re-use in the context of another story. Figure~\ref{fig-narrative_speech} shows an example of a generated narrative.

\begin{figure}
	\centering
	\begin{tabular}{|l|}
		\hline
		\underline{First} \textbf{I wanted} \textit{to} \textbf{get} \textit{the} \textbf{toy}.\\
		\underline{First} \textbf{you} \textbf{have} \textit{the} \textbf{toy}.\\
        \underline{Then} \textbf{I} \textbf{fail} \textit{to} \textbf{grasp it}.\\
        \underline{After} \textbf{I} \textbf{fail} \textit{to} \textbf{grasp}, \textbf{I} \textbf{reasoned}.\\
        \underline{Because} \textbf{I} \textbf{reasoned}, \textbf{I ask} \textit{for the} \textbf{toy} \textit{to} \textbf{you}\\
		\underline{Finally} \textbf{you gave me} \textit{the} \textbf{toy}\\
		\underline{Now} \textbf{I have} \textit{the} \textbf{toy}\\
		\textbf{I have} \textit{the} \textbf{toy} \underline{because} \textbf{you gave it} \textit{to} \textbf{me}\\
		\textbf{You gave} \textit{the} \textbf{toy} \textit{to} \textbf{me} \underline{because} \textbf{I wanted it}\\
		\hline
	\end{tabular}
	\caption{Example of a narrative generated by the robot. The \textit{language reservoir handler} will decompose the words in the narrative discourse in 3 categories: the \underline{discourse function words} (DFW) which direct the discourse from one sentence to the other, the \textbf{open class words} (OCW) which correspond to the meaningful words in terms of vocabulary of the sentence, and the \textit{closed class words} (CCW) which have a grammatical function in the sentence (see \cite{mealier2016}).
    }
     \label{fig-narrative_speech}
     
\end{figure}

\subsubsection{Show the learned kinematic structure}

As for the previous goals, this goal is generated through verbal requests. When asked ``What have you learned from your arm babbling?'', the iCub refers the human to look at the screen where the kinematic structures of its arms are displayed. Also, lines which connect nodes of the kinematic structures indicate the correspondences which the iCub has found between its left and right arm. Similarly, the iCub displays the correspondences which it has found between one of its arms and the body of the human (see Figure \ref{fig:kinematic-structure}). This knowledge is further employed to point to the human's arm, which is interesting as both the name as well as the kinematic location of the human's arm are inferred from self-learned representations and mapping these representations to the partner.

\subsection{Scenario Progression}
\label{sec:scenario_progression}

We first show how the full DAC-h3 system we have just described is able to generate a complex human-robot interaction by providing videos of live interactions (see \url{https://github.com/robotology/wysiwyd}) and a detailed description of a particular interaction. Then, in the next subsection, we will analyze more specifically the effect of the robot's proactivity level on naive users. In both cases, we consider a mixed-initiative scenario, where the iCub behaves autonomously as described in Section~\ref{sec:autonomous}, and so does the human. The human can interrupt the robot behavior by formulating verbal requests as described in Section~\ref{sec:goalexecution}. The scenario can follow various paths according to the interaction between the iCub's internal drive dynamics, its perception of the environment, and the behavior of the human. 

\begin{figure}[!t]
  \centering
  \includegraphics[width=0.46\textwidth]{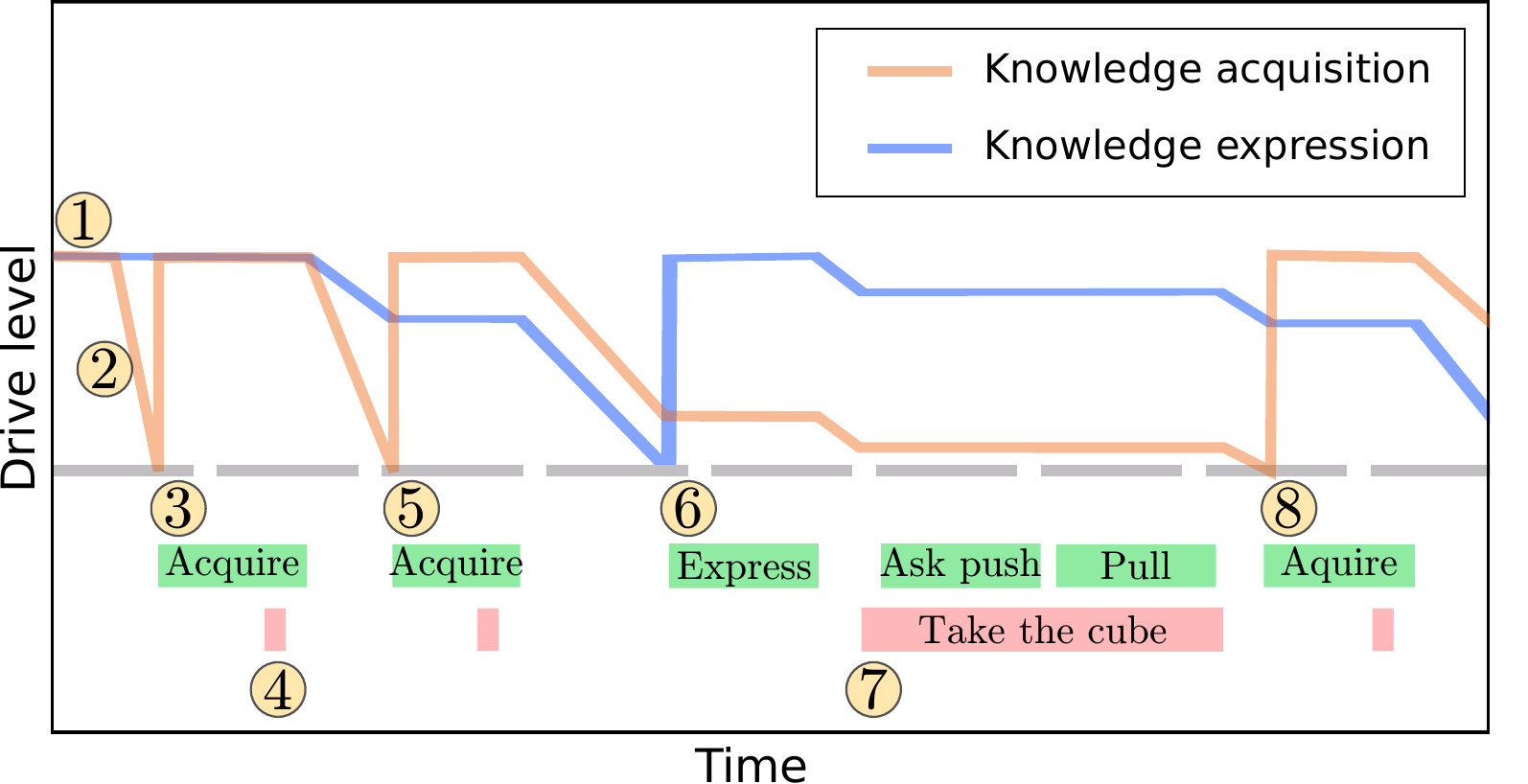}
 % \vspace{-0.25cm}
  \caption{Drive level dynamics during a typical mixed-initiative scenario described in Section~\ref{sec:scenario_progression}. Each drive starts at its default value and decays following the dynamics described in Section~\ref{sec:autonomous}. When reaching a given threshold (dashed horizontal line) the associated behavior is triggered (green rectangles), the corresponding drive level is reset to its default value and both drive levels are frozen for the duration of the behavior. Human behavior is indicated by the red rectangles, being either a reply to a question asked by the iCub (small rectangles) or a request to the iCub triggering goal oriented behavior (here: ``Take the cube''). The numbers refer to the description of the scenario progression in the main text. 
  }
 % \vspace{-0.25cm}
  \label{fig:drivedynamics}
\end{figure}

Here, we describe one particular instance of the scenario. Figure~\ref{fig:drivedynamics} shows the corresponding drive dynamics and human-robot interactions, and Figure~\ref{fig:wrdac} shows the connections between the modules of the cognitive architecture. Each of the numbered items below refers to its corresponding number in Figure~\ref{fig:drivedynamics}.

\begin{enumerate}
\item At the beginning of the interaction, the iCub has only limited knowledge about the current scene. In the \textit{sensations} module, the \textit{agent detector} detects the presence of a human and extracts its skeleton. The \textit{object recognition} module performs blob detection for extracting objects on the table from the visual input of the eye cameras and trains a classifier to categorize them. The 3D locations of the present objects are also computed in \textit{object recognition} through stereo vision. This forms a first incomplete representation of the scene in the \textit{episodic memory} where the \textit{object property collector} registers the location and type of each detected entity (here objects and an agent). It also contains slots for unknown body parts of the iCub, here the five fingers of its right hand. 

\item The presence of a large amount of missing information (presence of unknown objects, human and body parts) in the \textit{sensations} module makes the drive for knowledge acquisition to decay rapidly in the \textit{allostatic controller} and the drive for knowledge expression is kept constant (since there is no knowledge to express yet). 

\item When the knowledge acquisition drive level is below threshold, it triggers the associated behavior (\textit{behaviors} module) for acquiring information about an entity. The choice of the unknown entity is pseudo-random, with priority for requesting the name of an unknown human. This makes the robot look at the human. The visual input is passed to the \textit{perceptions} module where the \textit{synthetic sensory memory} segments the face from the background and attempts to recognize it from previously seen faces. If it does not recognize the face, the robot asks, ``I do not know you, who are you?''. The human can then reply, e.g., saying ``I am Daniel''. The level of the drive is reset to its default value and both drives are frozen during the behavior execution.

\item The perceived speech is analyzed by the \textit{language reservoir handler} in \textit{perceptions} to extract the name ``Daniel'' and is associated with the face representation in the \textit{associations} module. Thus, the next time the iCub will interact with this person, it will directly recognize him and not ask for his name.

\item Once this interaction is achieved, the drives continue to decay. Since the iCub has just acquired more information, the decay of the drive for knowledge acquisition is slower and the one for knowledge expression is increased. Still, the drive for knowledge acquisition reaches the threshold first. The behavior for acquiring information is therefore triggered again. This time, the random choice of an unknown entity makes the robot point to an object and ask, ``What is this object?''. The human replies e.g. ``This is the cube''. The \textit{language reservoir handler} extracts the name of the object from the reply and the \textit{associations} module associates it with the visual representation of the pointed object from \textit{object recognition}. Now the cube can later be referred by its name. 

\item The drives continue to decay. This time, the drive for knowledge expression reaches the threshold first. This triggers the behavior for expressing the acquired knowledge. A known entity is chosen, in this example it is the cube, which the robot points at while saying ``This is a cube''.

\item The human asks ``Take the cube''. A meaning is extracted by the \textit{language reservoir handler} in \textit{perceptions} and forms the representation of a goal to achieve in the \textit{goal} module (here the desired location of the object, i.e. the region of the iCub $I$ for the goal ``take'', see Figure~\ref{fig:state-transitions}). An action plan is built in \textit{action plans} with the sequence of two actions ``Ask the human to push the object'' then ``Pull the object'', together with their pre- and post-conditions in term of states ($I$, $S$ or $H$). The \textit{action selection} module takes care of the realization of the plan. First, it instantiates the action plan toward the considered object, here the cube, through its connection with \textit{associations}. Then, it executes each action until its associated post-condition is met (repeating it up to three times before giving up). Since the cube is in the human area $H$, the iCub first triggers the behavior for interacting verbally with the human, asking ``Can you please bring the cube closer to the shared area?''. The human pushes the cube to the shared area $S$ and the state transition is noticed by the robot thanks to the real-time object localization performed in the \textit{object recognition} module. Then the robot triggers a motor action to pull the cube. Once the goal is achieved (i.e. the cube is in $I$), the drive levels which were frozen during this interaction continue to decay.

\item The drive for knowledge acquisition reaches the threshold first. The associated behavior now chooses to acquire the name of a body part. The robot triggers the behavior for raising its hand and moving a random unknown body part, here the middle finger. It looks at the human and asks ``How do you call this part of my body?''. The name of the body part is extracted from the human's reply and is associated with the joint that was moved in \textit{associations}.

\end{enumerate}

The interaction continues following the dynamics of the drives and interrupted by the requests from the human. Once all available information about the present entities is acquired, the drive for knowledge acquisition stops to decay. However, the robot still maintains the interaction through its drive for knowledge expression and the human can still formulate requests for goal-oriented behavior. When new entities are introduced, e.g. an unknown object or another human entering the scene, the drive for knowledge acquisition decays again and the process continues. 

\subsection{Effect of the robot's proactivity level on naive users}

We now test the DAC-h3 architecture with naive users having to perform a collaborative task with the iCub robot. For this aim, we conduct experiments on six subjects and compare different configurations of the DAC-h3 drive system reflecting different levels of robot's proactivity.

To better control the experiment, we simplify the setup described above by limiting it to object tagging and manipulation. This means that in this study we do not use the functionalities for agent or body part tagging, action recognition, kinematic structure learning and narrative discourse. The iCub can only proactively tag or point at objects on the table, whereas the human can reply to object tagging requests and provide orders for taking or giving an object. These orders trigger action plans combining object manipulation by the iCub with requests to the human to do so, as described above. We note that, due to the distributed implementation of the DAC-h3 systems as interacting YARP modules operating in parallel, deactivating functionalities simply requires to not launch the corresponding modules and does not imply any modification in the code. 
%\cmf{here we could refer to the code paper, to see after acceptance}

\begin{figure}[!t]
  \centering
  \includegraphics[width=0.475\textwidth]{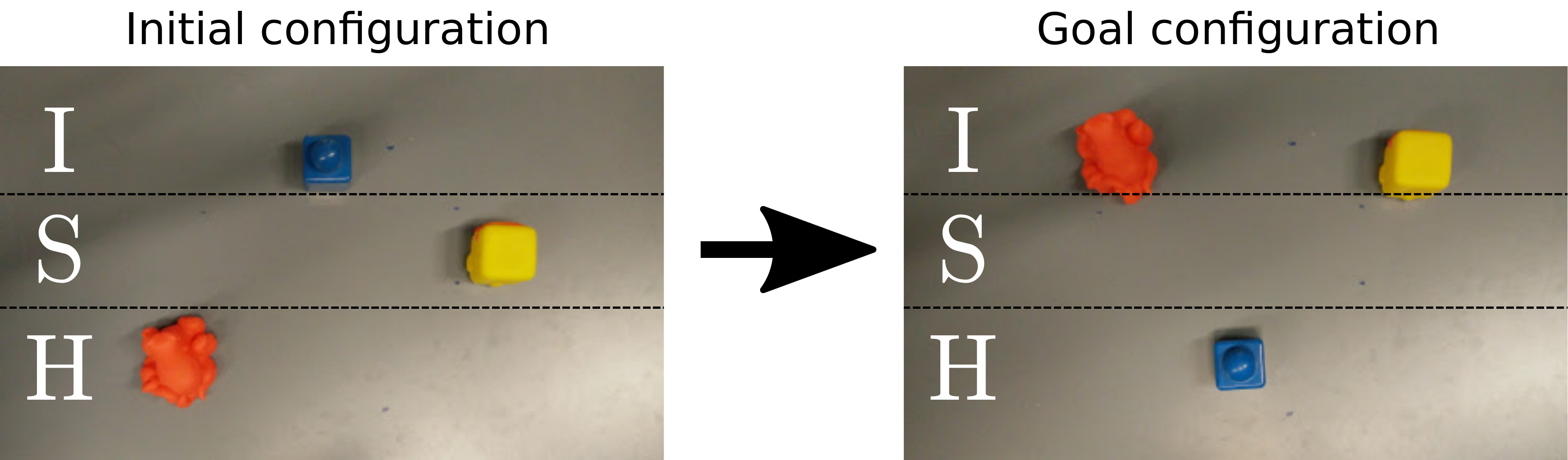}
  \caption{The experimental task. Each panel shows a top view of the table, where the three letters $I$, $S$ and $H$ indicates the iCub, Shared and Human areas as in Figure~\ref{fig:setup}.  Starting from the initial configuration of object positions on the left, the subject is asked to achieve the goal configuration on the right. This is done by interacting with the iCub following the instructions described above. The three objects are: an octopus (orange object on the left of the table), a blue cube (blue object in the middle) and a duck (yellow object on the right).
  }
  %\vspace{-0.25cm}
  \label{fig:init_goal_config}
\end{figure}

Three objects are placed on the table: an \textit{octopus}, a \textit{blue cube} and a \textit{duck}. Initially, the names of the objects are unknown to the iCub and they are placed on the table as shown on Figure~\ref{fig:init_goal_config} (left). The task given to the subjects is to achieve a goal configuration of object positions (Figure~\ref{fig:init_goal_config}, right). The experiment terminates when the task is achieved. 

To do so, the subject is instructed to interact with the iCub in the following way. At any time of the interaction, s/he can provide speech orders on the form of the sentences \textit{``Give me the <object name>''} and \textit{``Please take the <object name>''}, where \textit{<object name>} is the name of one of the three objects. These names are provided to the subject before the experiment starts to make the speech recognition more robust. Whenever the iCub asks the subject for the name of an object, s/he can reply with \textit{``This is the <object name>''}. Moreover, whenever the iCub asks to show a specific object, the subject can point to it using her/his right hand. This latter behavior is added to the state transition graph described in Figure~\ref{fig:state-transitions} and executed when the subject requests to perform an action on an object which is unknown to the iCub. To augment the difficulty of the task, the subject is asked to not move objects on its own initiative, but only when the iCub asks to do so.

Achieving the goal configuration requires a complex interaction between the subject and iCub. For example, moving the octopus from the human region to the iCub region requires to first inform the iCub about which object the octopus is, and then asking the iCub to take that object. Informing the robot about the name of an object can occur either from the iCub's initiative through the knowledge acquisition drive, or from the human's initiative by requesting an action on this object (when the object is unknown to the iCub, it will first ask the human to point at it). Since the octopus is not within the reach of the iCub, the robot will first ask the human to move it to the shared region, before executing the motor action for pulling the object in the iCub region. 

We run the experiment with six naive subjects on three different conditions. The three conditions correspond to three different levels of the robot's proactivity, defined by setting different drive decays in the \textit{allostatic controller} module. The two drives for knowledge acquisition and knowledge expression are initialized to a default value of 0.5. Then, they decay linearly at a rate of $n_{obj} * \delta$ unit/s, where $\delta$ is a constant value defining the proactivity level. For the knowledge acquisition drive, $n_{obj}$ is the number of perceived objects which are unknown to the robot. For the knowledge expression drive, it is the number of known objects. Therefore, the drive for knowledge acquisition (resp. knowledge expression) decays proportionally to the number of unknown objects (resp. known objects). We define three conditions: medium proactivity ($\delta$ for knowledge acquisition = 0.01, for knowledge expression = 0.004), slow proactivity ($\delta$ for both drives are 2.5 times lower than in medium proactivity) and fast proactivity ($\delta$ for both drives are 2.5 times higher than in medium proactivity). Corresponding behaviors are triggered when a drive value goes below 0.25. For example, for the knowledge acquisition drive in the medium-proactive condition ($\delta=0.01$) at the beginning of the interaction when all objects are unknown ($n_{obj}=3$), it takes approximately 8 seconds for the drive to decay completely from the default value of 0.5 to the threshold 0.25.

\begin{figure*}[!t]
  \centering
  \includegraphics[width=0.95\textwidth]{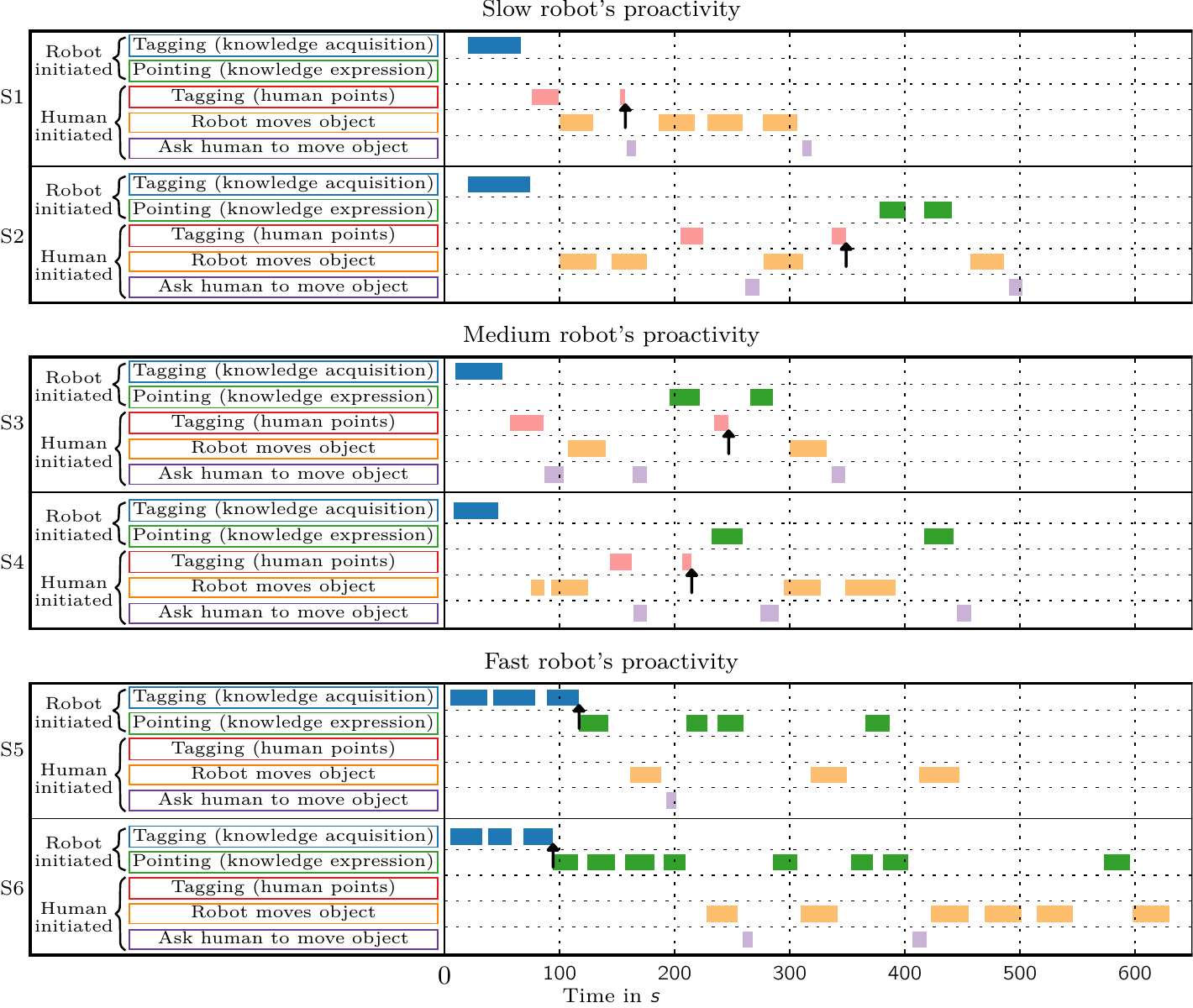}
  \caption{
  Interaction diagrams for the six subjects ($S1$ to $S6$, two per condition) interacting with the iCub to solve the task described in Figure~\ref{fig:init_goal_config}. Top: slow proactivity condition. Middle: medium proactivity. Bottom: fast proactivity. In each condition, the first five rows shows the interaction of one subject (e.g. $S3$ in the medium proactivity condition) and the five last rows the the interaction of another subject (e.g. $S4$ in the medium proactivity condition). For each subject, the colored bars show the times (x-axis) when specific actions (y-axis) are executed by the iCub. The two first rows (\textit{Tagging (knowledge acquisition)} and \textit{Pointing (knowledge expression)}) are actions initiated by the iCub through its drive regulation system in the allostatic controller as described in section~\ref{sec:autonomous}. The three last rows (\textit{Tagging (human points)}, \textit{Robot moves object} and \textit{Ask human to move object}) are actions initiated by the human through a speech order and executed by the iCub through action plans in the contextual layer, as described in section~\ref{sec:goalexecution}. They are sequenced according to the current state of the world and the current knowledge of the robot. The vertical arrow in each subject's interaction plot shows the time at which all the three object names are known to the robot. 
  }
 % \vspace{-0.25cm}
  \label{fig:interaction_diagrams}
\end{figure*}

Figure~\ref{fig:interaction_diagrams} shows the interaction diagrams for the six subjects grouped by condition. In the slow-proactive condition, we observe that the task can be solved very rapidly, as seen by the subject $S1$. Here the iCub has initiated only the first action (\textit{Tagging (knowledge acquisition)}). With drives decaying slower, the robot acts less often by its own initiative, which leads to the subject leading the interaction according to its own goal. Since the iCub is rarely acquiring information by itself, the subject has to request actions on objects which are not yet known to the robot, triggering tagging behaviors (\textit{Tagging (human points)}) prior to asking or moving objects. However, solving the task can also be quite long in this condition as seen by subject $S2$. This can be due to several factors such as the robot's failure to perceive or manipulate objects, or the time required by the subject to fully apprehend the system. This can also be due to the personality or the mood of the subject who can her/himself be more or less proactive. 

At the other extreme, in the fast proactive condition (bottom of the figure), we observe that the interaction is dominated by the iCub's self-generated actions (\textit{Tagging (knowledge acquisition)} and \textit{Pointing (knowledge expression)}). There is little time for the subject to request orders through speech between two consecutive actions of the iCub and it requires some time to succeed doing it. In consequence, solving the task can require quite a long time as seen by the subject $S6$. However, the names of all objects tend to be acquired faster in this condition (vertical arrows). This is because the drive for knowledge acquisition decays rapidly, especially at the beginning of the interaction where all objects are unknown, pushing the iCub to ask the name of everything around. Indeed, the three first actions are \textit{Tagging (knowledge acquisition)} for both $S5$ and $S6$.

Finally, in the medium-proactive condition, we observe a mixed initiative between the iCub and the subject. A possible positive effect of such a moderate proactivity of the robot is to maintain the social interaction when the subject is less proactive by her/himself, preventing long silent gaps in the interaction but still largely allowing the human to take the initiative. We observe a better interplay between iCub- and human-initiated actions in this condition.

Subjects are also asked to fill pre- and post-questionnaires before and after the interaction. The pre-questionnaire attempts at characterizing the personality traits of the subject, while the post-questionnaire evaluates subjective feelings about the interaction. As the main focus of this paper is on presenting a coherent cognitive architecture rather than the evaluation of the emerging human-robot interaction, we do not analyze these data in this paper. Providing a solid statistical analysis will require acquiring data of many more subjects which will be the aim of a follow-up paper. However, the presented human-robot interaction study shows that proactivity is an important factor in HRI and that it can be controlled by simple modifications of the drive dynamics. As observed in Figure~\ref{fig:interaction_diagrams}, the level of robot's proactivity has a clear influence on the resulting interaction by modulating the interplay between the robot and human initiatives.

%% file: 5_conclusion_futureworks.tex
\section{Conclusion and Future Works}
\label{sec:conclusion}
%\mh{It may be worth discussing which of the modules are "neural" and which are symbolic.  This was pointed out by Gregor Schoener, one of the reviewers. Perhaps this could be marked in one of the grand schematics. Also what is passed between them - there we can simply put YARP messages. In the discussion, we could mention that our architecture is currently hybrid (neural / symbolic) but we could gradually replace everything by something more biologically motivated in the future...} \cmf{I have done it a bit below, to check. But we don't have that much neural modules in the current version actually, so the distinction is more about purely algorithmic implementations (e.g. planner and ABM) vs. more biologically plausible models (e.g. ... SAM?)}

This paper has introduced \emph{DAC-h3}, a proactive robot cognitive architecture to acquire and express knowledge about the world and the self. The architecture is based on the Distributed Adaptive Control (DAC) theory of the brain and mind, which provides a biologically grounded framework for organizing various functional modules into a coherent cognitive architecture. Those modules implement state-of-the-art algorithms modeling various cognitive functions for autonomous self-regulation, whole-body motor control, multimodal perception, knowledge representation, natural language processing, and goal-oriented behavior. They are all implemented using the YARP robotic middleware on the iCub robot, ensuring their parallel execution in real time and providing synchronous and asynchronous communication protocols among modules.

The implementation of DAC-h3 is flexible so that existing modules can easily be replaced by more sophisticated ones in the future. Moreover, most modules can be configured according to the user's needs, for example by adding new drives or more complex grammars into the system. This makes DAC-h3 a general framework for designing autonomous robots, especially in HRI setups. The underlying open-source code contains some modules which are specific to the iCub robot (modules related to action execution; which work on both real and simulated iCub's), but all cognitive nodes (autobiographical memory, allostatic controllers, etc.) can be freely used with other robots (most easily with YARP-driven robots, but thanks to YARP-ROS intercommunication also with ROS driven robots). 
%Researchers can benefit from this architecture to study Human-Robot Interactions during learning and proactive experiments.

The main contribution of this paper is not about the modeling of the specific functional modules, which already have been published (see Section~\ref{sec:wrdac}), but rather about the integration of a heterogeneous collection of modules into a coherent and operational cognitive architecture. For this aim, the \emph{DAC-h3} architecture is organized as a layered control structure with tight coupling within and between layers (Figure~\ref{fig:wrdac} and Section~\ref{sec:wrdac}): the \emph{somatic}, \emph{reactive}, \emph{adaptive}, and \emph{contextual} layers. Across these layers, a columnar organization exists that deals with the processing of states of the world or exteroception, the self or interoception, and action. Two main control loops generate the behavior of the robot. First, a reactive-adaptive control loop ensures autonomy and proactivity through the self-regulation of internal drives for knowledge acquisition and expression. It allows the robot to proactively manage its own knowledge acquisition process and to maintain the interaction with a human partner, while associating multimodal information about entities with their linguistic labels. Second, an adaptive-contextual control loop allows the robot to satisfy human requests, triggering goal-oriented behavior relying on the acquired knowledge. Those goal-oriented behaviors are related to action planning for object passing, pointing, action recognition, narrative expression and kinematic structure learning demonstration. 

%The DAC-h3 architecture provides a principled methodology for organizing functional cognitive modules into a well-defined architecture. 
We have shown that these two control loops lead to a well-defined interplay between robot-initiated and human-initiated behaviors, which allows the robot to acquire multimodal representations of entities and link them with linguistic symbols, as well as to use the acquired knowledge for goal-oriented behavior.
This allows the robot to learn reactively as well as proactively. Reactive learning occurs in situations where the robot requires obtaining new knowledge to execute a human order (e.g. grasping an object with an unknown label), and thus leads to an efficient interaction for acquiring the information before acting according to the human desire.
At the same time, the robot can also learn proactively to optimize its cognitive development by triggering learning interactions itself, which allows the robot to learn without having to wait for the human to teach new concepts. Moreover, this is supposed to reduce the cognitive load of the human teacher, as the robot 1) chooses the target entity of the learning interaction, 2) engages the joint attention by making the human aware of the target entity, and 3) asks for the corresponding label. Therefore, the human only needs to provide the label, without having to be concerned about the prior knowledge of the robot. 

We have implemented the entire \emph{DAC-h3} architecture and presented an HRI scenario where an iCub humanoid robot interacts with objects and a human to acquire information about the present objects and agents as well as its own body parts. We have analyzed a typical interaction in detail, showing how \emph{DAC-h3} is able to dynamically balance the knowledge acquisition and expression processes according to the properties of the environment, and to deal with a mixed initiative scenario where both the robot and the human are behaving autonomously. In a series of video recordings, we show the ability of \emph{DAC-h3} to adapt to different situations and environments. We have also conducted experiments with naive subjects on a simplified version of the scenario, showing how the robot's proactivity level influences the interaction.

Adapting the proactivity level also provides a step towards personalities of robots. The curiousness (i.e. favoring proactive learning) or talkativeness (i.e. communicating about its own knowledge) of the robot is determined by the decay rates of the corresponding drives. Thus the personality of the robot can be altered by a simple modification of the decay values, as done for skill refinement by Puigbo et al.~\cite{puigbo2015}.

The current work has the following limitations. First, some of the available abilities deserve to be better integrated into the HRI scenario. For example, this is the case for the kinematic structure learning process which is currently executed in a separated learning phase instead of being fully integrated within the interaction scenario. Similarly, the narrative can only be generated from specific chunks of the robot's history as recorded in the autobiographical memory. Second, in this paper, we do not provide a statistical analysis of the HRI experiments. The reason is that we focus on the description of the entire architecture and on their theoretical principles. A thorough statistical analysis will require collecting much more data to fully demonstrate the utility of some of these principles, for example how proactivity helps to solve the referential indeterminacy problem, as well as the effect of the robot's autonomy on the acceptability of the system by naive users. Third, although \emph{DAC-h3} can solve parts of the \emph{Symbol Grounding Problem} (SGP), it still presupposes a symbolic concept of an entity which is given \textit{a priori} to the system. Therefore, our contribution is more about the ability to proactively acquire multimodal information about these entities and linking them to linguistic labels that can be reused to express complex goal-oriented behavior later on.

We are currently extending the proposed architecture in the following ways. First, we are better integrating some of the available abilities within the interaction scenario as mentioned above. This will allow starting the knowledge acquisition process from scratch in a fully autonomous way. Second, we are considering to use more biologically plausible and/or computationally scalable models for some of the existing modules, namely the \emph{action planning} and \emph{action selection} modules. These are currently algorithmic implementations using predefined action plans. We want to replace it with an existing model of rule learning grounded in the neurobiology of the prefrontal cortex which can learn optimal action policies from experience to maximize long-term reward~\cite{Duff2011}. An interesting feature of this model for solving the SGP is that it relies on neural memory-units encoding sensorimotor contingencies with causal relationships learned through adaptive connections between them. An alternative solution is to use state-of-the art AI planners such as the Planning Domain Definition Language (PDDL), where multi-agent planning extensions are particularly relevant in the context of social robotics \cite{kovacs2012multi}.  %Such sensorimotor contingencies, adaptive ly learned from experience, could provide the basis of emerging symbols without relying on a pre-existing concept of entity as in the current version of the architecture. 
Third, we are also integrating more low-level reactive control abilities through an acquired notion of a peri-personal space \cite{Roncone2016peripersonal}, where the robot will be able to optimize its own action primitives to maintain safety distances with aversive objects (e.g. a spider) in real time while executing reaching actions toward other objects. Finally, we are working on a self-exploration process to autonomously discover the area which is reachable by the robot, similarly to Jamone et al.~\cite{jamone2014}, and subsequently employing this self-model and applying it to the human partner to estimate his/her reachability. %\tf{Self learning of reachable areas to add}